\pdfoutput=1
\documentclass{article}
\usepackage{preprint,times}

\usepackage{amsmath,amsfonts,bm}

\def\eqref#1{equation~\ref{#1}}

\def\1{\bm{1}}

\def\va{{\bm{a}}}

\def\vt{{\bm{t}}}

\def\vw{{\bm{w}}}
\def\vx{{\bm{x}}}

\def\vz{{\bm{z}}}

\DeclareMathAlphabet{\mathsfit}{\encodingdefault}{\sfdefault}{m}{sl}
\SetMathAlphabet{\mathsfit}{bold}{\encodingdefault}{\sfdefault}{bx}{n}

\newcommand{\E}{\mathbb{E}}

\usepackage{graphicx}
\graphicspath{{figures/}}
\usepackage{xcolor}
\usepackage{booktabs}
\usepackage{array}
\usepackage{enumitem}
\setlist[itemize]{leftmargin=*,itemsep=1pt,topsep=2pt,parsep=0pt}
\setlist[enumerate]{leftmargin=*,itemsep=1pt,topsep=2pt,parsep=0pt}
\usepackage{hyperref}
\hypersetup{colorlinks=true,linkcolor=blue!50!black,citecolor=blue!50!black,urlcolor=blue!50!black}
\usepackage{url}
\newcommand{\sr}[2]{#1$_{\pm#2}$}
\newcommand{\pmr}[2]{#1 $\pm$ #2}
\newcommand{\best}[1]{\textbf{#1}}

\newcommand{\na}{\textcolor{gray}{--}}

\newcommand{\rone}{$r_1$}
\newcommand{\dg}{$^{\circ}$}
\newcommand{\res}{$^{\ast}$}
\newcommand{\apptab}[1]{Appendix Table~\ref{#1}}
\newcommand{\apptabs}[2]{Appendix Tables~\ref{#1} and~\ref{#2}}
\newcommand{\appfig}[1]{Appendix Figure~\ref{#1}}

\newcommand{\pttwo}{PT$^2$-LLM}
\newcommand{\dtwo}{D$^2$Quant}
\newcommand{\quipsharp}{QuIP\#}

\title{RoleQ: Role-Aligned Scales for Converting Language Models to Ternary}

\author{Johannes Knittel\\
Harvard University\\
\And
Alec Helbling\\
Georgia Tech \\
\And
Hanspeter Pfister\\
Harvard University \\
}

\begin{document}

\maketitle

\begin{abstract}
Weight quantization below four bits shrinks memory and bandwidth at a perplexity that looks acceptable, but often at the cost of factual knowledge and task accuracy. Ternary weights have mostly been obtained by training from scratch or long fine-tuning.
We introduce RoleQ, a quantization framework that places scales by the functional roles of Transformer units rather than by the layout of the weight matrix, as conventional formats do.
Vectors that read from the residual stream take one scale per output unit; those that write to it, one per input unit. The rule is structural, needs no search, calibration or stored state, and holds on integer and ternary grids.
In a rotated basis it matches a per-layer axis search at half the solver work, and at ternary it cuts perplexity by up to half against layout-based placement in two families at every size from 0.6B to 14B, where per-row scales leave MMLU at chance.
Our lightweight post-training pipeline converts an existing checkpoint to 1.59 bits per weight in a few GPU-hours, a fraction of the cost of learned ternarization, and leads every post-training ternary method at matched or lower memory-resident bits on perplexity, zero-shot accuracy and MMLU, most widely on knowledge. A sparse second ternary plane placed by a per-unit Fisher saliency keeps the matmul multiplication-free and recovers what ternary loses first: at 8B, a fraction of a bit returns about half the MMLU that conversion costs, well ahead of random or Hessian-based placement at equal bits.
\end{abstract}

\section{Introduction}

Weight memory and bandwidth constrain language-model deployment on phones and in browsers. At batch size one, decoding is typically bandwidth-bound, making bits per weight a key determinant of latency~\citep{wang2025bitnetcpp,wei2025tmac}, motivating ultra-low-bit quantization. Below four bits, however, acceptable perplexity can conceal severe losses in factual knowledge and task accuracy. We therefore evaluate quantization with three complementary measures: WikiText2 perplexity, a seven-task zero-shot average and five-shot MMLU.

Ternary weights replace weight multiplications with additions and subtractions. Pretraining~\citep{wang2023bitnet,ma2024bitnet158}, quantization-aware training~\citep{liu2025paretoq}, code fine-tuning~\citep{malinovskii2024pvtuning} and learned ternarization over many epochs~\citep{wang2026catq,wang2026ayot} preserve much of the model's knowledge at substantial computational cost. Post-training conversion~\citep{zhang2026qronos,yan2026pt2llm,guo2026qtea} is much cheaper but loses factual knowledge: on Qwen3-8B-Base, these methods score 26--35 on MMLU, versus 77 at full precision. The challenge is to approach training-based retention at post-training conversion cost.

Our central insight is that quantization should allocate bits by function. We introduce RoleQ, a format that assigns scales to functional units: vectors that read from the residual stream receive one scale per output unit; vectors that write to it receive one per input unit. This structural rule applies to integer and ternary grids and requires neither calibration data nor stored axis choices. RoleQ-T combines this format with a data-free learned rotation, ternary GPTQ~\citep{frantar2023gptq} with an exact per-vector step, and brief scale distillation with frozen codes (Figure~\ref{fig:pipeline}). Qwen3-8B-Base~\citep{yang2025qwen3} reaches 1.59 bits per weight (bpw), 11.33 perplexity, 60.3 zero-shot accuracy and 47.6 MMLU in a few GPU-hours, a fraction of the tens of A100-hours for CAT-Q's learned ternarization at this size (Section~\ref{sec:exp-recipe}).

Functional units also guide the allocation of extra bits. At a tight budget, selecting units by quantization-error magnitude can perform worse than random; per-unit Fisher saliency estimates how strongly each unit's error affects the loss. RoleQ-T therefore optionally adds a sparse second ternary plane to Fisher-selected units, preserving multiplication-free matrix products. At 2.16 bpw, Qwen3-8B-Base reaches 9.30 perplexity, 65.9 zero-shot accuracy and 62.8 MMLU. On post-trained Qwen3-8B, 1.99 bpw surpasses CAT-Q on all three measures for 0.28 additional bits.

We make three contributions.
\begin{enumerate}
\item \textbf{Role-aligned scale placement} (Section~\ref{sec:roleq}). In a rotated basis, RoleQ matches AdaDim's~\citep{heo2024adadim} reconstruction quality at half the solver work of a two-way search, without calibration data or stored axis choices. Per-vector INT4 matches MXFP4 at 8\% fewer bits; ternary perplexity improves over the best uniform placement by up to $2\times$, with gains at every tested size from 0.6B to 14B across two families. At 14B, RoleQ raises MMLU from 25.4 (chance) with per-output-channel scales to 38.5.
\item \textbf{Efficient ternary conversion} (Sections~\ref{sec:rotations}--\ref{sec:sd}). RoleQ-T outperforms Qronos, \pttwo{} and QTEA on all three measures at matched or lower memory-resident rates, with the largest gains on knowledge. At 14B, it surpasses QTEA's released checkpoint after scale distillation using 0.67 fewer bpw. Applying the same distillation to \pttwo{} leaves it over four perplexity points behind at a higher rate.
\item \textbf{Fisher-guided secondary planes} (Section~\ref{sec:planes}). At 8B, +0.57 bpw recovers about half the MMLU lost on Base. On the post-trained model at 1.99 bpw, Fisher selection beats random by 6.2 points and Hessian$\times$weight by 9.0.
\end{enumerate}

\begin{figure}[t]
  \centering
  \includegraphics[width=\textwidth]{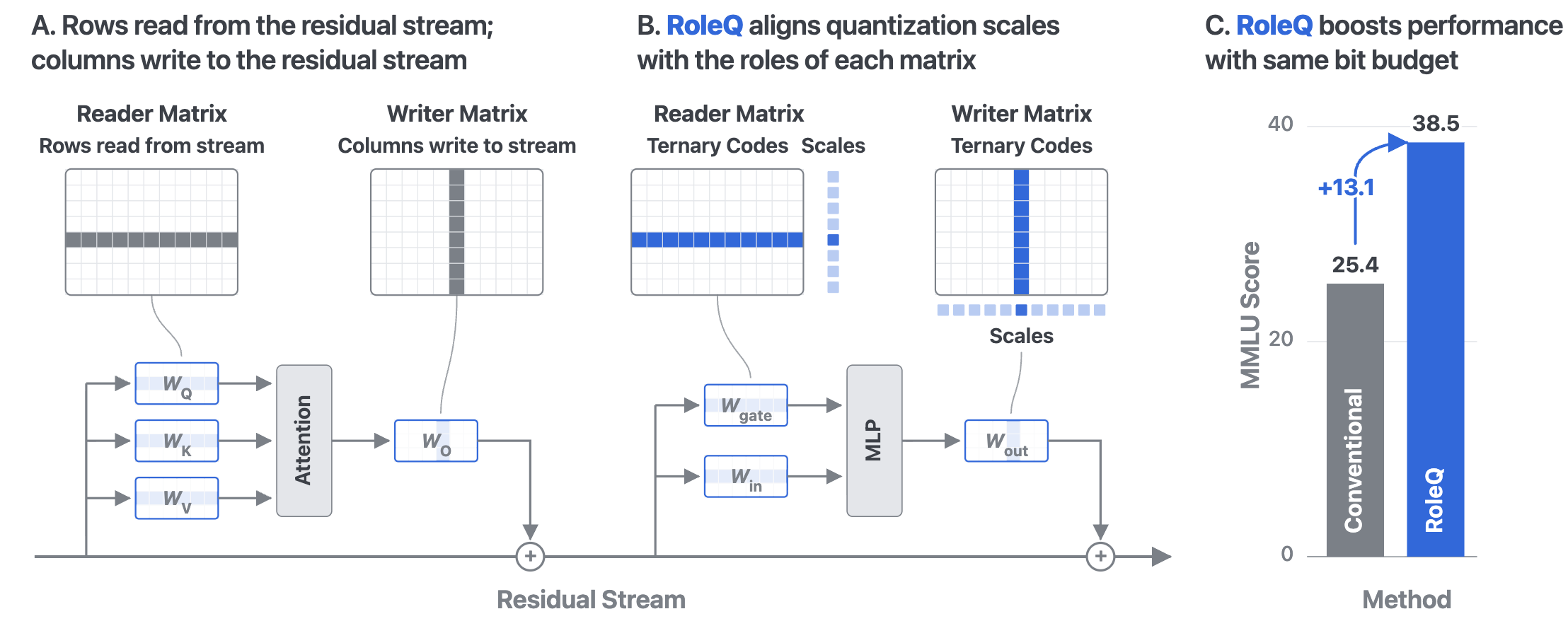}
  \caption{\textbf{RoleQ aligns scales with the roles.} (A) Readers read from the residual stream through rows, writers write to it through columns. (B) RoleQ places the scales on the \emph{unit} axis, per row for readers and per column for writers, never on the stream axis. (C) Qwen3-14B, ternary at equal bits: the placement alone lifts MMLU from 25.4 (chance) with per-output-channel scales to 38.5.}
  \label{fig:layouts}
\end{figure}

\section{Related Work}

\textbf{Ternary weights and conversion.} Ternary weights $\{-1, 0, +1\}$ date at least to \citet{hwang2014fixedpoint}; ternary weight networks~\citep{li2016twn} popularised the scaled ternary code with a threshold rule. Following binary BitNet~\citep{wang2023bitnet}, BitNet b1.58~\citep{ma2024bitnet158} showed that language models pretrained with this alphabet can match full-precision counterparts at scale while enabling multiplication-free matrix products. Converting existing checkpoints avoids pretraining from scratch, but the computational cost varies. Quantization-aware training, distillation and code fine-tuning~\citep{liu2025paretoq,wu2025bitnetdistillation,malinovskii2024pvtuning} still update weights or codes. Learned ternarization also involves sustained optimization: CAT-Q~\citep{wang2026catq} uses 512 calibration samples but optimises its codes for 60 epochs, while ScaleQ-1.58~\citep{wang2026ayot} uses the model's own reasoning traces. Although CAT-Q is described as post-training quantization, we distinguish its extended code optimization from the lightweight conversion setting considered here. Direct post-training methods include Qronos~\citep{zhang2026qronos}, \pttwo{}~\citep{yan2026pt2llm}, QTEA~\citep{guo2026qtea}, PTQTP~\citep{xiao2025ptqtp} and TWLA~\citep{zhao2026twla}. We compare these methods at memory-resident rates (\apptabs{tab:competitors}{tab:bits}), since several report sub-2-bit code counts that exclude strictly necessary decoding metadata. TWLA's released implementation requires 4.15--4.65 bpw, so it serves as a reference rather than a bit-matched baseline. PTQTP adds a second plane to every unit, while QTEA adds FP8 residuals to curvature-selected columns; neither allocates these extra bits using a per-unit loss-sensitivity score.

\textbf{Two-bit formats.} \quipsharp{}~\citep{tseng2024quipsharp} and QTIP~\citep{tseng2024qtip} use lattice and trellis codebooks, respectively, achieving strong two-bit results on Llama. GuidedQuant's loss-guided Hessian~\citep{kim2025guidedquant} and YAQA's end-to-end adaptive rounding~\citep{tseng2026yaqa} build on QTIP's trellis codes. QTIP adopts \quipsharp{}-style fine-tuning, reported at 50--100 GPU-hours for a 70B model. These methods provide strong quality references. Our format is scalar and codebook-free, and its ternary codes also support multiplication-free matrix products. For a comparison within this format class, our equal-bits controls reproduce the uniform grouped grids used by GPTQ~\citep{frantar2023gptq} and EfficientQAT~\citep{chen2025efficientqat}.

\textbf{Scale axis and rotation.} Existing methods choose scale axes empirically or refine scales within a solver. AdaDim~\citep{heo2024adadim} selects per-input or per-output scales for each module through a reconstruction test without rotation, and reports lower quality when per-input scales are applied uniformly rather than selectively. ShiftAddLLM~\citep{you2024shiftaddllm} refits column scales during the solver's sweep; Z-FOLD~\citep{jeon2023zfold} adds a folded in-channel scale on a second axis, and \dtwo{}~\citep{yan2026d2quant} adds one for a single matrix type. We adopt Z-FOLD's closed-form $\mathbf H$-metric refit for readers. These methods use calibration data and do not derive scale placement from vector roles. RoleQ instead selects the axis structurally from each vector's reader or writer role, without calibration or stored axis flags. Our rotations retain the slots and parametrization of QuaRot~\citep{ashkboos2024quarot} and SpinQuant~\citep{liu2025spinquant}, using SpinQuant's notation. OptRot~\citep{gadhikar2025optrot} and TWLA learn rotations without data, whereas HARP~\citep{zagitov2026harp} learns a stored rotation from calibration data before codebook quantization. Our rotation contribution is the choice of objective, evaluated against those of OptRot and TWLA.

\textbf{Saliency and mixed allocation.} BiLLM~\citep{huang2024billm} and ARB-LLM~\citep{li2025arbllm} assign residual binary planes to columns selected by curvature scores. FGMP~\citep{hooper2025fgmp} allocates precision using $\sum g^2\delta^2$ with a single global percentile, an allocator we adopt at unit granularity. MixLLM~\citep{zheng2026mixllm}, SqueezeLLM~\citep{kim2024squeezellm} and TaCQ~\citep{xiao2025tacq} use gradient, Fisher and task-conditioned scores. We condition our score on factual recall and find that, at tight budgets, curvature-based selection can perform worse than random placement.

\section{Method}

\begin{figure}[t]
  \centering
  \includegraphics[width=0.92\textwidth]{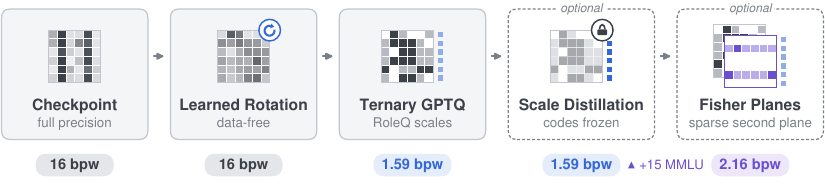}
  \caption{\textbf{The RoleQ-T pipeline.} A data-free learned rotation precedes ternary GPTQ with an exact per-vector step, giving codes in $\{-1, 0, +1\}$ (grey) and one scale per functional unit (blue). The dashed stages are optional: scale distillation refits scales with frozen codes, and Fisher planes add a sparse second plane to salient units (+0.57 bits here). Values below each stage are memory-resident bits per weight for Qwen3-8B-Base.}
  \label{fig:pipeline}
\end{figure}

RoleQ places quantization scales by the functional roles of weight vectors on both integer and ternary grids (Section~\ref{sec:roleq}). RoleQ-T (Sections~\ref{sec:rotations}--\ref{sec:sd}) converts an existing checkpoint to this ternary format in a few GPU-hours, combining a data-free learned rotation, GPTQ adapted to role-aligned scales and brief scale distillation with frozen codes. Optional Fisher-placed secondary planes (Section~\ref{sec:planes}) allocate additional precision to units selected by loss sensitivity. Figure~\ref{fig:pipeline} summarises the pipeline. The placement rule is data-free, the solver and distillation use unlabeled data.

\paragraph{Preliminaries.}
A weight is reconstructed as $w \approx s\,(q - z)$ from an $L$-level code $q$, a scale $s$ and an optional zero point $z$, shared by a group of weights. Standard placement uses one scale per output channel (per-OC), i.e.\ per row of an $[\mathrm{out}, \mathrm{in}]$ matrix; per-input-channel (per-IC) scales index columns. Ternary codes represent $\{-1, 0, +1\}\cdot s$ at $\log_2 3 = 1.585$ bpw before scale overhead; rates strictly between this and 2 bpw require non-power-of-two alphabets or mixed rates (Section~\ref{sec:planes}). Round-to-nearest (RTN) fits grids from weights alone. Compensated solvers~\citep{frantar2022obc,frantar2023gptq} estimate the input second moment $\mathbf{H} = \E[\vx\vx^\top]$ from unlabelled documents and propagate rounding errors to unquantized coordinates. Rates count $\log_2 L$ bits per code plus all strictly necessary decoding metadata (resident bpw, Appendix~\ref{app:bits}).

\subsection{RoleQ: Role-Aligned Scales}
\label{sec:roleq}

RoleQ assigns one scale to each vector that \emph{reads} from the residual stream (a row of $W_Q$, $W_K$, $W_V$, $W_{\mathrm{gate}}$, $W_{\mathrm{in}}$) or \emph{writes} to it (a column of $W_O$, $W_{\mathrm{out}}$). This rule follows a mechanistic view of the Transformer: the residual stream is a shared communication channel, and components access it through dense read and write vectors. An MLP neuron has read vectors in $W_{\mathrm{in}}$ and $W_{\mathrm{gate}}$ and a write vector in $W_{\mathrm{out}}$, corresponding to the \emph{key} and \emph{value} of \citet{geva2021keyvalue}. Attention reads through Q/K/V rows and writes through O columns, forming the QK and OV circuits of \citet{elhage2021mathematical}. These vectors interact through the stream~\citep{knittel2026sparsedep}; each has a magnitude, direction and functional role. Assigning each vector its own scale separates its magnitude from the direction approximated by its quantization codes. In contrast, a scale per residual coordinate (per-OC on writers, per-IC on readers) couples unrelated units through a shared coordinate. RoleQ therefore aligns the quantization grid with the network's functional decomposition. In a rotated residual basis, this distinction remains intact: the rotation mixes stream coordinates while preserving the unit axis (Section~\ref{sec:rotations}).

Standard per-OC placement suits row-major storage and W8A8 accumulators. Weight-only decoding also permits input-axis scales, applied to activations once per token at $O(d)$ cost for input width $d$, though per-IC placement has remained rare. Prior methods choose the axis by layout or search rather than function. RoleQ uses one rule on integer and ternary grids, \emph{readers scale per output unit, writers per input unit} (Figure~\ref{fig:layouts}), with no search, calibration pass or stored per-layer flag. Writer scales lie on the solver's sweep axis, requiring the adaptation in Section~\ref{sec:solver}. At inference, they simply rescale the input activations.

\subsection{Rotations}
\label{sec:rotations}

RoleQ-T rotates weights before quantization using QuaRot's rotation slots in SpinQuant's notation~\citep{ashkboos2024quarot,liu2025spinquant}: $r_1$ on the residual basis, $r_2$ mixing V$\to$O per key/value head, and $r_4$ on the MLP intermediate dimension (Appendix~\ref{app:rotations}). Rotations reduce the impact of outliers on low-bit quantization~\citep{chee2023quip,ashkboos2024quarot,liu2025spinquant}: a random Hadamard alone lowers 1.7B ternary PTQ perplexity from 1245.9 to 119.8. We therefore use a rotated residual basis throughout our quantization runs, placement controls and rerun baselines; only the basis control includes unrotated models. Our contribution concerns the rotation objective. We learn $s_1$ and $s_2$ in the $r_1$ and $r_2$ slots from weights alone, without calibration, and compare three objectives: \emph{opt}, the exact ternary reconstruction error $\sum_{\vw}\min_{a,\vt}\|\vw-a\vt\|^2 = \sum_{\vw}\big(\|\vw\|^2 - \max_S(\sum_{i\in S}|w_i|)^2/|S|\big)$ over rotated reader and writer vectors (Eq.~\ref{eq:ternary}), differentiable almost everywhere; \emph{l4}, OptRot's fourth-power ($\ell_4$) objective, a kurtosis proxy~\citep{gadhikar2025optrot}; and \emph{trigmm}, TWLA's tri-modal shaping~\citep{zhao2026twla}. For $s_1$, \emph{opt} achieves the lowest weight-space error but worse perplexity than \emph{l4} (130.3 versus 89.99 at 1.7B; Appendix~\ref{app:ablations}). The recipe uses $s_1(\mathrm{l4}) + s_2(\mathrm{opt}) + r_4$, reducing 8B perplexity by a further factor of $1.38$ relative to random Hadamard rotation. These rotations add 0 bpw and fold into the weights, apart from $r_4$'s runtime Hadamard. The placement rule is defined in this rotated basis: $r_1$ mixes residual-stream coordinates and leaves RoleQ's scale axis untouched.

\subsection{Ternary GPTQ: the RoleQ-T Solver}
\label{sec:solver}

RoleQ-T adapts GPTQ~\citep{frantar2023gptq} to approximately minimise the reconstruction objective $\mathrm{tr}[(\mathbf W-\hat{\mathbf W})\mathbf H(\mathbf W-\hat{\mathbf W})^\top]$, where $\hat{\mathbf W}$ reconstructs $\mathbf W$ and $\mathbf H$ is its input second moment. GPTQ sweeps input coordinates, quantizes each column and propagates its error to the remaining columns. We adapt the per-vector ternary fit and its position within this sweep (derivations and settings in Appendix~\ref{app:recipe}). Both reader and writer kernels use an exact vector fit, but at different stages.

For a nonzero vector $\vw$, scale $a$, ternary codes $\vt$ and nonempty support $S$ (the nonzero codes), the joint optimum of $\min_{a,\vt}\|\vw-a\vt\|^2$ is
\begin{equation}
\vt=\mathrm{sign}(\vw)\odot[\,|\vw|>\tau\,],\qquad
a^\star(S)=\frac{1}{|S|}\sum_{i\in S}|w_i|,\qquad
S^\star=\arg\max_{S}\frac{\big(\sum_{i\in S}|w_i|\big)^2}{|S|},
\label{eq:ternary}
\end{equation}
where $\odot$ denotes elementwise multiplication, $[\cdot]$ the indicator, and $S$ ranges over nonempty prefixes of $|\vw|$ sorted in descending order. The threshold $\tau$ selects $S^\star$~\citep{yin2019quantization,hou2018lossaware}; TWN instead fixes $\tau = 0.7\cdot\mathrm{mean}|w|$~\citep{li2016twn}. The corresponding exact grid search handles integer grids and the four-level INT2-g128 control, replacing the min/max placement that LeanQuant and NeUQI replace by a search over candidate grids~\citep{zhang2025leanquant,lin2026neuqi}. All placements in Table~\ref{tab:placement} thus share one code path.

With $\mathbf W\in\mathbb R^{m\times d}$ in $[\mathrm{out},\mathrm{in}]$ layout, codes $\mathbf T$ and scales $\va$ (rows for readers, columns for writers),
\begin{equation}
\hat{\mathbf W} = \mathrm{diag}(\va)\,\mathbf T \ \ \text{(readers)}, \qquad
\hat{\mathbf W} = \mathbf T\,\mathrm{diag}(\va) \ \ \text{(writers)},
\label{eq:format}
\end{equation}
so writer scales lie on the sweep axis and reader scales on the orthogonal axis. At writer column $j$, GPTQ has compensated the entire vector for previous rounding errors. Equation~\ref{eq:ternary} jointly fits its scale $a_j$ and codes $\vt_j$, then GPTQ propagates the error to subsequent columns. For readers, each step touches one coordinate per row; re-searching a row's support would revisit already quantized entries. We therefore initialise each row's grid $\{-a_i, 0, a_i\}$ using Eq.~\ref{eq:ternary} and hold it fixed during the sweep. Since the objective separates across rows, we then refit each scale at the final codes as $a_i^\star = \vt_i^\top\mathbf H\vw_i/\vt_i^\top\mathbf H\vt_i$. This refit cannot increase the reconstruction error and preserves per-unit ternary weights. Exactness applies to the vector fits and refits; the overall sweep remains greedy.

\subsection{Scale Distillation}
\label{sec:sd}

Scale distillation (+SD) refits only the scales, parametrised as $\log a$, and norm gains, 0.02--0.08\% of the parameters depending on model size and plane budget (0.024\% for the 8B body). It minimises KL divergence from the full-precision teacher's next-token distribution while keeping codes frozen, preserving the per-unit ternary format. Before quantization, rotation folds RMSNorm's per-coordinate gains into adjacent weights. SD reintroduces a trainable gain vector at each norm site (two per block and the final norm, +0.0007 bpw), initialised to one, allowing the student to rescale residual coordinates. We use only unlabelled data: 400 steps at batch size 8, or 3,200 windows of 512 tokens from RedPajama-1T-Sample. The teacher's final residuals are cached once during quantization and mapped to logits by its frozen output head, avoiding teacher forward passes during SD. At 8B, the stage takes 2\,h on one A100 (Appendix~\ref{app:recipe}), against 8--64 A100-hours for CAT-Q's 60-epoch stage at this size and 160 for ScaleQ-1.58's~\citep{wang2026catq,wang2026ayot}; \quipsharp{}-style fine-tuning, which QTIP adopts, is reported at about 100 A100-hours for a 70B model~\citep{tseng2024quipsharp,tseng2024qtip}, and PV-Tuning reports days on eight GPUs at 70B~\citep{malinovskii2024pvtuning}. Applying the same SD to a baseline in Section~\ref{sec:exp-recipe} tests whether distillation alone explains the recipe's advantage.

\subsection{Fisher-Placed Sparse Secondary Planes}
\label{sec:planes}

Sparse secondary planes target the task accuracy and factual knowledge lost in ternary conversion. Existing methods add precision more broadly or on a fixed axis. PTQTP~\citep{xiao2025ptqtp} gives every weight a second ternary plane. TWLA's released code~\citep{zhao2026twla}, which matches its reported perplexity and task numbers, adds a second plane to 26\% of weights, unmentioned in the paper; and QTEA~\citep{guo2026qtea} adds FP8 residuals to salient columns, the latter two using layer-local saliency. We instead select whole \emph{functional units} by loss sensitivity: the scaled vectors of Eq.~\ref{eq:format}, in every matrix the rows of a reader or the columns of a writer, i.e., neurons and attention dimensions that read from or write to the residual stream. Each selected unit receives a second ternary code and scale on the same axis, $\hat{\vw} = a_1\vt_1 + a_2\vt_2$, fitted by the same GPTQ to the residual $\vw - a_1\vt_1$. This represents up to nine levels with $2\log_2 3$ code bits per weight and evaluates as two ternary matmuls and an addition (Appendix~\ref{app:planes}).

We rank units by a Fisher-weighted saliency, a second-order proxy for the loss increase from one-plane quantization~\citep{LeCun1989OptimalBD}, with the diagonal empirical Fisher in place of the loss Hessian as in Fisher pruning and SqueezeLLM~\citep{theis2018fisher,kim2024squeezellm}. It weights errors by their effect on the target loss, whereas GPTQ's $\mathbf H$ measures layer-local reconstruction error.
Let $\mathcal L_n$ be the mean per-token negative log-likelihood of item $n$'s target tokens (a factual answer or the second half of a text window) given the preceding tokens. Define $F_i = \frac{1}{N}\sum_n (\partial\mathcal L_n/\partial w_i)^2$ for weight $w_i$ in the rotated full-precision model.
The saliency of unit $u$ is
\begin{equation}
S_u = \sum_{i\in u} F_i\,\delta_i^2, \qquad \boldsymbol\delta = \vw - a^\star\vt,
\label{eq:saliency}
\end{equation}
where $\boldsymbol\delta$ is the residual of the exact one-plane fit in Eq.~\ref{eq:ternary}, which the second plane targets. The score uses observed-target gradients as a sensitivity proxy, not an exact estimate of recoverable loss. We rank units of both roles across all layers and matrix kinds by $S_u$ and give the top $K$ a second plane, testing this common ranking in Section~\ref{sec:exp-planes}.

The number $K$ of selected units sets the rate in unit-sized steps from 1.590 to 3.179 bpw (two planes everywhere) with the same kernel; we set it from a target of 1.994 bpw for post-trained models and 2.156 for Base (the INT2-g128 rate), or from a baseline's rate.

The items used to accumulate the Fisher determine its sensitivity. C4 calibration windows yield a \emph{text Fisher}; 720 factual-recall items (MMLU-style multiple choice, TriviaQA cloze and ARC; \citealp{hendrycks2021mmlu,joshi2017triviaqa,clark2018arc}), rendered through the chat template and drawn with a seed disjoint from our in-house probe, yield a \emph{knowledge Fisher}. The ranking selects units, not weights, so these items decide where a plane goes but never what it stores. We independently vary the item set and the scoring checkpoint (the Base model or its post-trained sibling). Ranking agreement across two disjoint item halves measures stability: Spearman correlations range from 0.87 to 0.98 (Appendix~\ref{app:planes}).

\section{Experiments}

\label{sec:setup}

\textbf{Models and data.} We quantize Qwen3-0.6B, 1.7B, 8B and 14B Base and post-trained Qwen3-1.7B and 8B~\citep{yang2025qwen3}, with Llama-3.1-8B-Base~\citep{grattafiori2024llama3} as a cross-family check and Llama-3.1-8B-Instruct as the matched-bits comparison with QTIP's and GuidedQuant's released checkpoints (\apptab{tab:llama}). Base calibration uses 512 unlabelled C4 sequences of 512 tokens~\citep{raffel2020t5}.
Post-trained models use 272 chat sequences of up to 1536 tokens, half UltraChat conversations~\citep{ding2023ultrachat} and half self-generated responses.

\textbf{Evaluation.} We report WikiText2 perplexity, the average of seven zero-shot tasks (PIQA, ARC-Easy, ARC-Challenge, HellaSwag, WinoGrande, BoolQ and OpenBookQA; \citealp{bisk2020piqa,clark2018arc,zellers2019hellaswag,sakaguchi2020winogrande,clark2019boolq,mihaylov2018openbookqa}) under lm-eval~\citep{gao2024lmeval}, and five-shot MMLU~\citep{hendrycks2021mmlu}. These measures capture complementary effects: modest perplexity changes can accompany substantial losses in factual knowledge~\citep{jaiswal2024compressing}, which perplexity measures only coarsely and quantization degrades on its own axis~\citep{namburi2023cost,wang2026compressedlens}, while matching aggregate scores can conceal disagreement on individual predictions~\citep{dutta2024accuracy}.

\textbf{Comparison protocol.} We report memory-resident information bits, including decoding metadata, name rate and stage differences, evaluate baseline reruns and released checkpoints in our harness against published results, and treat our seed spreads as descriptive.
Further details are given in Appendix~\ref{app:setup}.

\subsection{Ternary Conversion Recipe}
\label{sec:exp-recipe}

We first ask what quality a few GPU-hours of ternary conversion can retain. Table~\ref{tab:ternary} compares RoleQ-T with ternary methods whose outputs we could evaluate and memory-resident rates establish (Qronos, \pttwo{}, QTEA, CAT-Q and TWLA; \citealp{zhang2026qronos,yan2026pt2llm,guo2026qtea,wang2026catq,zhao2026twla}), using each baseline's best available stage. On Qwen3-8B-Base, RoleQ-T reaches 1.590 bpw with scale distillation in three to four A100 hours, and 2.157 bpw with probing and planes in under ten. Four-level (INT2) grids provide a separate equal-bits comparison in Sections~\ref{sec:exp-axis} and~\ref{sec:exp-planes}.

At 1.590 bpw, RoleQ-T after scale distillation leads Qronos, \pttwo{} and QTEA on all three measures while using matched or fewer bits: the baselines require 1.591, 1.835 and 2.281 bpw, respectively. Adding planes improves all three scores further, still below QTEA's rate. Knowledge retention distinguishes these methods particularly clearly. Without our distillation, the baselines score 25.9--34.8 on MMLU, close to or modestly above 25\% chance accuracy and far below 76.8 at full precision. RoleQ-T retains 47.6 at 1.590 bpw and 62.8 at 2.157 (56.0 before distillation). Perplexity alone therefore gives an incomplete account of the knowledge lost during conversion.

Two comparisons delimit this result. First, shared scale distillation does not close the gap: applying it to \pttwo{}'s codes yields 15.51 perplexity at 1.835 bpw, versus 11.33 for RoleQ-T at 1.590 (Table~\ref{tab:ternary}). The advantage thus survives the same distillation procedure, although the quantization recipes differ. Second, comparison with learned ternarization entails a budget trade-off. CAT-Q optimises its codes for 60 epochs on 512 C4 samples, 8--64 A100-hours at 8B by its own bracket (Appendix~\ref{app:bits}), and its released checkpoint scores 52.9 MMLU at 1.71 bpw on post-trained Qwen3-8B on our harness. Our planes reach 61.8 at 1.99 bpw in under ten GPU-hours: +8.9 points for 0.28 additional bits. At CAT-Q's own 1.71 bpw, the corresponding RoleQ-T-g128 leads it on perplexity (17.00 vs 27.68) and narrowly on zero-shot (61.7 vs 61.0) and ties on MMLU (53.1 vs 52.9).

The gains extend to 14B. With planes and scale distillation, RoleQ-T reaches 8.05 perplexity ($1.26\times$ full precision), 69.2 zero-shot and 68.3 MMLU at 2.156 bpw, versus 11.78, 56.6 and 47.8 for QTEA's released checkpoint at 2.26. At the matched PTQ stage, before distillation, the planes run still leads on all three measures (10.11, 65.7 and 62.0), with each gap at least $19\times$ our seed range. This includes WikiText2, QTEA's calibration corpus. The one-plane run at 1.589 bpw also leads after distillation (9.53 / 65.4 / 59.7), saving 0.67 bpw, but trails at PTQ. The lower-rate advantage therefore depends on distillation; the planes advantage holds at both stages. TWLA~\citep{zhao2026twla} is a higher-rate reference: it releases no checkpoint, and its released code, which reproduces its reported Qwen3-8B perplexity, builds a masked two-plane model its paper does not describe, resident at 4.15--4.65 bpw, roughly twice our planes rate (Table~\ref{tab:ternary}, Appendix~\ref{app:bits}).
Section~\ref{sec:limitations} discusses the stronger trellis-coded results at matched bits on Llama-3.1-8B-Instruct (\apptab{tab:llama}).

\begin{table}[t]
\caption{\textbf{RoleQ-T leads the tested post-training ternary baselines below 2.3 bpw.} WikiText2 perplexity, seven-task zero-shot average and five-shot MMLU; baselines at their best available stage, ours at PTQ and after scale distillation (+SD), mean of two seeds $\pm$ half-range. TWLA (higher rate) and CAT-Q (learned) are references. Planes use the text Fisher on Base models and the knowledge Fisher on the post-trained model.
}
\label{tab:ternary}
\begin{center}
\footnotesize
\setlength{\tabcolsep}{2.5pt}
\begin{tabular}{@{}llrrrrr@{}}
\toprule
Method & Stage & claimed bpw & resident bpw & WT2 ppl$\,\downarrow$ & Zero-shot$\,\uparrow$ & MMLU$\,\uparrow$ \\
\midrule
\multicolumn{7}{@{}l}{\textbf{Qwen3-8B-Base}\quad fp16: 6.99 / 70.42 / 76.82} \\
\multicolumn{7}{@{}l}{\textit{one ternary plane}} \\
Qronos + HIP + MagR & PTQ & 1.58 & 1.591 & 22.80 & 46.67 & 25.93 \\
\pttwo{} & PTQ & 1.58 & 1.835 & 67.95 & 44.88 & 26.52 \\
\pttwo{} codes + our SD & +SD & 1.58 & 1.835 & 15.51 & 56.37 & 36.74 \\
RoleQ-T (ours) & PTQ & 1.585 & 1.590 & \pmr{22.42}{0.22} & \pmr{50.39}{0.09} & \pmr{33.32}{1.48} \\
RoleQ-T (ours) & +SD & 1.585 & 1.590 & \best{\pmr{11.33}{0.06}} & \best{\pmr{60.33}{0.27}} & \best{\pmr{47.59}{2.72}} \\
\multicolumn{7}{@{}l}{\textit{a second (sparse) plane}} \\
QTEA & PTQ & 1.70 & 2.281 & 15.66 & 50.60 & 34.82 \\
RoleQ-T + planes (ours) & PTQ & 2.156 & 2.157 & \pmr{11.32}{0.01} & \pmr{59.51}{0.30} & \pmr{56.04}{0.03} \\
RoleQ-T + planes (ours) & +SD & 2.156 & 2.157 & \best{\pmr{9.30}{0.02}} & \best{\pmr{65.89}{0.34}} & \best{\pmr{62.75}{0.41}} \\
\midrule
\textit{TWLA}, released code & PTQ & 1.58 & 4.15--4.65 & 9.03 & 65.11 & 62.72 \\
\midrule
\multicolumn{7}{@{}l}{\textbf{Qwen3-8B post-trained}\quad fp16: 9.72 / 69.36 / 74.93} \\
CAT-Q (learned, 60 epochs) & learned & 1.58 & 1.710 & 27.68 & 61.03 & 52.92 \\
RoleQ-T, one plane (ours) & +SD & 1.585 & 1.590 & \pmr{19.59}{1.58} & \pmr{59.96}{0.32} & \pmr{48.41}{1.46} \\
RoleQ-T-g128 (ours) & +SD & 1.710 & 1.711 & \pmr{17.00}{0.20} & \pmr{61.67}{0.03} & \pmr{53.10}{0.15} \\
RoleQ-T + planes (ours) & +SD & 1.994 & 1.990 & \best{\pmr{15.30}{0.36}} & \best{\pmr{64.57}{0.28}} & \best{\pmr{61.82}{0.15}} \\
\midrule
\textit{TWLA} & PTQ & 1.58 & 4.15--4.65 & 12.56 & 62.97 & 62.78 \\

\midrule
\multicolumn{7}{@{}l}{\textbf{Qwen3-14B-Base}\quad fp16: 6.38 / 73.39 / 80.66} \\
RoleQ-T (ours) & PTQ & 1.585 & 1.589 & \pmr{18.10}{0.42} & \pmr{55.50}{1.11} & \pmr{44.72}{0.41} \\
RoleQ-T (ours) & +SD & 1.585 & 1.589 & \best{\pmr{9.53}{0.05}} & \best{\pmr{65.37}{0.03}} & \best{\pmr{59.75}{0.20}} \\
\midrule
QTEA, released checkpoint & PTQ & 1.70 & 2.257 & 11.78 & 56.62 & 47.75 \\
RoleQ-T + planes (ours) & PTQ & 2.156 & 2.156 & \pmr{10.11}{0.04} & \pmr{65.69}{0.20} & \pmr{62.00}{0.34} \\
RoleQ-T + planes (ours) & +SD & 2.156 & 2.156 & \best{\pmr{8.05}{0.04}} & \best{\pmr{69.21}{0.01}} & \best{\pmr{68.31}{0.15}} \\
\bottomrule
\end{tabular}
\end{center}
\end{table}

\subsection{The Placement Rule}
\label{sec:exp-axis}

Table~\ref{tab:placement} isolates scale placement: within each grid and model, the placements share the solver, random Hadamard $r_1$, calibration draw and paired seeds, with no distillation.
At ternary, RoleQ has the lowest mean perplexity at every tested size in both families, improving on the best uniform placement by up to $2\times$, with the same ordering on every seed from 1.7B upward.
At 14B, the gain also appears on both task measures: 52.3 versus 45.6 zero-shot and 38.5 versus chance-level 25.4 MMLU for per-OC (\apptab{tab:axis-seeds}). The 0.6B and Qwen-8B perplexity gaps remain within the largest seed range.
Integer grids show the same mean perplexity ordering, but the margin depends on alphabet size and model.
On the four-level grid, the best uniform placement has $1.19\times$ RoleQ's perplexity at 1.7B and $1.22\times$ on Llama-8B, versus about $1.01\times$ at Qwen-8B and 14B.
At INT3 and INT4, the largest gains occur at 1.7B: ratios of $2.19\times$ and $1.14\times$, respectively.
Placement thus matters most at ternary, where the per-unit scale is a unit's only continuous degree of freedom; here the benefit persists to the largest tested model in each family.
Two practical comparisons follow. RoleQ-INT4 matches or improves on MXFP4~\citep{ocp2023mx,rouhani2023microscaling} perplexity at 0.6B, 1.7B and 8B while using 8\% fewer bits (information bits, Appendix~\ref{app:bits}).
The rule also comes within 0.023\% of AdaDim's~\citep{heo2024adadim} searched $\mathbf H$-metric reconstruction error at half the solver work of a two-way GPTQ search, without calibration data for axis selection or stored axis flags. This is a reconstruction comparison, not an end-to-end AdaDim evaluation. Appendix~\ref{app:axis} controls for codes, scale count and fitting procedure, and shows that the rule's advantage depends on rotation.

\begin{table}[t]
  \centering
  \footnotesize
  \setlength{\tabcolsep}{3pt}
  \caption{\textbf{Role-aligned scale placement achieves the lowest perplexity over fixed per-channel placement.}
   WikiText2 perplexity at four Qwen3-Base sizes and Llama-3.1-8B-Base, under matched PTQ settings (random Hadamard \rone{}). Mean of two ($^{4}$ four) seeds $\pm$ half-range; lower is better.
  }
  \label{tab:placement}
  \vspace{0.3em}
  \begin{tabular*}{\textwidth}{@{\extracolsep{\fill}} l c c c c c c @{}}
    \toprule
     & & \multicolumn{4}{c}{Qwen3-Base} & \\
    \cmidrule(lr){3-6}
    Configuration & bpw & 0.6B & 1.7B & 8B & 14B & Llama-8B \\
    \midrule
    fp16 & 16.000 & 12.66 & 9.39 & 6.99 & 6.38 & 6.30 \\
    \midrule
    Ternary: per-OC & 1.590 & \pmr{387.2}{34.0} & \pmr{222.3}{8.0} & \pmr{35.47}{0.52} & \pmr{37.89}{0.65} & \pmr{114.4}{8.6} \\
    Ternary: per-IC & 1.590 & \pmr{1187}{226} & \pmr{305.0}{19.8} & \pmr{33.39}{1.70}$^{4}$ & \pmr{42.60}{0.48} & \pmr{669.9}{28.1} \\
    Ternary: RoleQ (ours) & 1.590 & \best{\pmr{351.9}{21.8}} & \best{\pmr{136.2}{8.8}} & \best{\pmr{31.28}{0.77}}$^{4}$ & \best{\pmr{25.25}{0.34}} & \best{\pmr{58.13}{2.29}} \\
    \midrule
    INT2-g128: per-OC & 2.156 & \pmr{79.87}{1.65} & \pmr{37.87}{0.45} & \pmr{13.11}{0.08}$^{4}$ & \pmr{10.90}{0.05} & \pmr{49.7}{14.0} \\
    INT2-g128: per-IC & 2.156 & \pmr{139.1}{9.7} & \pmr{41.43}{1.30} & \pmr{13.83}{0.10} & \pmr{11.64}{0.05} & \pmr{23.28}{0.81} \\
    INT2-g128: RoleQ (ours) & 2.156 & \pmr{77.55}{0.67} & \pmr{31.79}{0.58} & \pmr{12.93}{0.12} & \best{\pmr{10.76}{0.01}} & \pmr{19.15}{0.09} \\
    Ternary + planes (ours) & 2.157 & \best{\pmr{53.96}{1.43}} & \best{\pmr{29.41}{0.92}} & \best{\pmr{12.83}{0.05}} & \pmr{12.19}{0.14} & \best{\pmr{15.39}{0.08}} \\
    \midrule
    INT3: per-OC & 2.810 & \pmr{19.89}{0.09} & \pmr{33.76}{0.21} & \pmr{7.97}{0.02} & \pmr{7.17}{0.01} & \pmr{8.29}{0.02} \\
    INT3: per-IC & 2.810 & \pmr{25.52}{0.57} & \pmr{25.98}{0.38} & \pmr{8.17}{0.01} & \pmr{8.08}{0.01} & \pmr{90.9}{6.6} \\
    INT3: RoleQ (ours) & 2.810 & \best{\pmr{19.24}{0.04}} & \best{\pmr{11.87}{0.00}} & \best{\pmr{7.95}{0.02}} & \best{\pmr{7.16}{0.01}} & \best{\pmr{8.12}{0.03}} \\
    \midrule
    INT4: per-OC & 3.910 & \pmr{13.99}{0.03} & \pmr{11.29}{0.04} & \pmr{7.24}{0.03} & \pmr{6.56}{0.01} & \pmr{6.72}{0.00} \\
    INT4: per-IC & 3.910 & \pmr{15.16}{0.04} & \pmr{11.81}{0.05} & \pmr{7.25}{0.01} & \pmr{6.67}{0.00} & \pmr{7.55}{0.01} \\
    INT4: RoleQ (ours) & 3.910 & \best{\pmr{13.86}{0.04}} & \best{\pmr{9.90}{0.02}} & \pmr{7.20}{0.03} & \best{\pmr{6.53}{0.01}} & \best{\pmr{6.70}{0.00}} \\
    MXFP4: block of 32 & 4.250 & \pmr{14.12}{0.01} & \pmr{10.10}{0.00} & \best{\pmr{7.19}{0.01}} & \pmr{6.54}{0.01} & \pmr{6.79}{0.00} \\
    \bottomrule
  \end{tabular*}
\end{table}

\subsection{Sparse Secondary Planes}
\label{sec:exp-planes}

Secondary planes recover much of the knowledge lost in single-plane conversion for a fraction of a bit. At 8B, +0.57 bpw of text-Fisher planes recovers about half the MMLU loss: scores rise from 33.3 to 56.0 at PTQ and from 47.6 to 62.8 after scale distillation, against 76.8 at full precision. After distillation, perplexity improves from 11.33 to 9.30 (\apptab{tab:recipe-ladder}). At 14B, planes raise MMLU from 44.7 to 62.0 at PTQ and from 59.7 to 68.3 after distillation, with zero-shot gains of 10.2 and 3.8 points. All six 14B gains exceed four seed ranges, but five are smaller than their 8B counterparts (Table~\ref{tab:ternary}), suggesting greater benefit where single-plane conversion loses more quality.
Placement determines how much this budget recovers (Table~\ref{tab:saliency}). At 1.99 bpw on post-trained Qwen3-8B, knowledge-Fisher selection beats random placement by 6.2 MMLU points and Hessian$\times$weight by 9.0; the latter falls below random. On post-trained 1.7B, the full selector set separates by score family (\apptab{tab:selectors}): loss-sensitivity scores yield 9.7--10.3 PTQ perplexity, versus 13.7--17.9 for error-magnitude scores, with random placement in the latter range. The 1.7B budget sweep supports this distinction: random placement's perplexity relative to Fisher falls from $1.96\times$ at $+0.15$ bits to $1.12\times$ at $+0.9$ (\appfig{fig:sweep}).

At matched bits, the comparison with four-level grids depends on size and basis (Table~\ref{tab:placement}; \apptab{tab:equal-bits}). Under $r_1$, standard per-OC INT2-g128 has $1.48\times$ and $1.29\times$ the planes run's perplexity at 0.6B and 1.7B. In the recipe basis, the corresponding ratios are $1.039\times$ at Qwen-8B and $1.27\times$ at Llama-8B PTQ. Planes also outperform the stronger role-placed grid at 0.6B, at Qwen-8B PTQ and on Llama at both stages. At 14B, however, four-level grids lead under $r_1$ (10.76 and 10.90 versus 12.19), so neither format dominates across sizes. On task accuracy, Llama's PTQ gains over per-OC exceed the seed spread (+2.9 zero-shot, +10.7 MMLU). Qwen's PTQ task differences have split signs and lie within one seed range; after distillation, the MMLU gain is +2.4 points but remains below three ranges. Ternary planes thus offer competitive quality at equal bits while retaining a multiplication-free matmul with no codebook or group metadata.

\begin{table}[t]
  \centering
  \small
  \setlength{\tabcolsep}{4.5pt}
  \caption{\textbf{Our Fisher selector outperforms random and Hessian-based selection at matched bit rates.}
  Results on post-trained Qwen3-8B after scale distillation (two seeds, $\pm$ half-range).}
  \label{tab:saliency}
  \vspace{0.4em}
  \begin{tabular*}{\textwidth}{@{\extracolsep{\fill}} l c c c c @{}}
      \toprule
      Configuration / Score
        & Total bpw
        & WikiText-2 PPL $\downarrow$
        & 7-Task Zero-Shot Avg. $\uparrow$
        & MMLU $\uparrow$ \\
      \midrule
      One Plane
        & 1.590
        & $19.59 \pm 1.58$
        & $59.96 \pm 0.32$
        & $48.41 \pm 1.46$ \\
      Hessian $\times$ Weight
        & 1.988
        & $17.68 \pm 0.14$
        & $62.32 \pm 0.19$
        & $52.80 \pm 0.17$ \\
      Random Units
        & 1.991
        & $15.74 \pm 0.18$
        & $63.07 \pm 0.08$
        & $55.58 \pm 0.03$ \\
      Knowledge Fisher (Ours)
        & 1.990
        & $\mathbf{15.30 \pm 0.36}$
        & $\mathbf{64.57 \pm 0.28}$
        & $\mathbf{61.82 \pm 0.15}$ \\
      \midrule
      Full Precision (FP16)
        & 16.000
        & 9.72
        & 69.36
        & 74.93 \\
      \bottomrule
  \end{tabular*}
\end{table}

\section{Conclusion}
\label{sec:conclusion}
\label{sec:limitations}

We introduced RoleQ, which assigns quantization scales to functional units, and RoleQ-T, which converts existing checkpoints to ternary in a few GPU-hours. Two findings stand out. First, the Transformer's division of labour into readers and writers provides a structural rule for scale placement. In a rotated basis, this rule requires no axis search, calibration data or stored axis choices, and its benefits persist to the largest models tested. Second, post-training conversion can retain useful task accuracy below two bits per weight. RoleQ-T outperforms the tested ternary conversion baselines at matched or lower memory-resident rates, with the largest gains in factual knowledge, which perplexity alone does not capture. Fisher-guided secondary planes further recover knowledge by directing a fraction of a bit to loss-sensitive units. Together, these results offer a practical route to lower memory use and multiplication-free ternary matrix products while preserving useful model capabilities.

These findings have a defined scope. The placement rule is supported in a rotated basis; its advantage does not extend to native coordinates. Four Qwen3 sizes and one Llama model do not establish a general scaling trend, and the best uniform placement varies with size and family. Fisher selection's advantage over random placement narrows with increasing budget in the 1.7B sweep. The gains from adding planes are also generally smaller at 14B than at 8B at the same total rate. At 2 bpw on Llama, QTIP's trellis codes, alone and with GuidedQuant's Hessian~\citep{tseng2024qtip,kim2025guidedquant}, remain ahead (\apptab{tab:llama}), so we do not claim parity with this codebook class. Our evaluations do not cover generation or reasoning, including long chains of thought, where quantization introduces distinct failure modes~\citep{liu2025quantizationreasoning}. Our results are weight-only (A16) and measure storage and quality, not inference speed. The ternary inner loop requires no multiplier, codebook or group metadata, making evaluation with lookup-table kernels~\citep{wang2025bitnetcpp,wei2025tmac} a natural next step (Appendix~\ref{app:bits}).

The broader lesson concerns allocation. At very low precision, both scale sharing and the placement of additional bits shape knowledge retention. Under tight budgets, allocating precision to the right units can matter more than increasing the overall rate. A low-bit format should therefore be judged by both how many bits it spends and where it spends them.

\subsubsection*{Acknowledgments}
This work was partially supported by the Harvard Data Science Initiative Postdoctoral Fellowship.

\bibliography{references}
\bibliographystyle{preprint}

\appendix

\section{Nomenclature and Bit Accounting}
\label{app:bits}

\paragraph{Nomenclature.}
INT$b$ denotes a grid with $2^b$ levels; our symmetric per-vector grids INT3sym / INT4sym / INT8sym have 7 / 15 / 255 levels, so INT4sym costs $\log_2 15$ bits plus one fp16 scale per vector, 3.91 bpw (4.00 in storage bits, so its 8\% saving against MXFP4's 4.25 in Section~\ref{sec:exp-axis} is 6\% in storage bits). We write \emph{ternary} for the three-level code and reserve INT2 for four-level grids, namely our exact-grid control and the external 2-bit methods. Rates are stated in information bits ($\log_2 L$ per code) so that a ternary total compares directly with an integer one. We quantize weights only (W$\cdot$A16): activations, embeddings and the head stay in fp16, as in every method we compare against (Qwen3-8B: 13.9\,GB bf16 body, 1.38\,GB at 1.59 bpw, 2.49\,GB of fp16 embeddings and head).

\paragraph{Scale granularity.}
Scales are shared per tensor, per output channel (one per row of $[\mathrm{out}, \mathrm{in}]$), per (row, group of $g$) or per input channel (one per column). Side information amortises over the group as $b + (16 + b_z)/g$ bpw: INT2-g128 with fp16 scale and zero point is 2.156, MXFP4 is $4 + 8/32 = 4.25$~\citep{ocp2023mx,rouhani2023microscaling}, INT4-g128 is 4.125, and one fp16 scale per vector is $16/d$, which vanishes (0.004 at $d = 4096$). Every dense matrix has a stream axis ($d_{\mathrm{model}}$) and a unit axis (heads $\times$ $d_{\mathrm{head}}$, or $d_{\mathrm{ff}}$): five kinds read from the residual stream through their rows ($W_Q$, $W_K$, $W_V$, $W_{\mathrm{gate}}$, $W_{\mathrm{in}}$) and two write to it through their columns ($W_O$, $W_{\mathrm{out}}$), so per-OC is the unit axis on readers and the stream axis on writers, and per-IC the reverse.

\paragraph{Resident terms.}
The information rate counts $\log_2 L$ per code plus every term a decoder needs to produce the weight: scales, zero points, act-order indices, group or plane membership, per-group affine terms, rotations that are neither folded into the weights nor regenerable from a seed, and reintroduced norm gains. Folded or seed-regenerable rotations cost 0 bpw. Packing (5 trits per byte = 1.6, 2 bits per trit = 2.0) can only add to the information rate and is not charged. We apply one test to every claimed rate, our own included: can a decoder produce the weight from the counted bits alone? By this test the per-vector scales and reintroduced norm gains move RoleQ-T from 1.585 to 1.590 bpw; TWLA's per-(row, group) $\alpha$ and $\mu$ and its mask membership are consumed at inference, which places it at 4.15--4.65; and QTEA's 1.7-bit payload is resident at 2.28. Table~\ref{tab:bits} gives the arithmetic term by term.

\paragraph{Stages.}
We distinguish five stages of increasing cost: calibration-free PTQ; calibrated PTQ (GPTQ); \emph{scale distillation}, which re-fits the scales and norm gains to the fp model's next-token distribution with the codes frozen (0.02--0.08\% of the parameters, 400 steps); fine-tuning of codes or weights with the quantizer in the loop~\citep{malinovskii2024pvtuning,wang2026catq,tseng2024qtip}, from tens of A100-hours at 8B (CAT-Q) to about 160 (ScaleQ-1.58) and about 100 at 70B for \quipsharp{}-style fine-tuning (Section~\ref{sec:sd}); and training from scratch~\citep{wang2023bitnet}. Tables~\ref{tab:ternary} and~\ref{tab:competitors} record the stage of every run, and every comparison in the paper is either stage-aligned or names both stages.

\paragraph{The hardware case.}
Ternary weights offer two savings rather than one. (i) Memory traffic scales with bpw; it is identical for every format at equal bits and is the saving today's GPUs realise. (ii) An integer weight at any width still multiplies (dequantize-and-FMA or an integer MAC, plus a scale and zero point every group), whereas a ternary weight adds the activation, subtracts it or skips it: no multiplier, no codebook and no group metadata in the inner loop. The per-vector scale costs one multiply per output or input coordinate, $O(d)$ per token against $O(d^2)$ additions, and a second plane costs two accumulations and an add. Binary weights share this property but have no zero; four-level and wider grids do not have it. The second saving is realised today on CPUs by lookup-table kernels~\citep{wang2024bitnetcpp,wang2025bitnetcpp,wei2025tmac} and is the premise of add-only hardware, for which BitNet b1.58 estimates about $70\times$ lower matmul arithmetic energy than fp16 at 7\,nm~\citep{ma2024bitnet158}. We report no latency or energy measurements of our own (Section~\ref{sec:limitations}).

\begin{table}[h]
\caption{\textbf{Compared methods.} Each method's claimed and memory-resident rate (derivations in Table~\ref{tab:bits}), whether it trains codes or weights, whether it calibrates on data, where its scales sit, and how it enters our tables (rerun: its own code on our checkpoint; ckpt: its released checkpoint decoded and scored on our harness; paper: cited as reported). PTQTP has no public code or checkpoints; its rate and scale axis follow from its own decomposition and storage formula~\citep{xiao2025ptqtp}. QTIP, alone and under GuidedQuant's loss-guided Hessian, is the trellis-coded 2-bit class we do not claim to match; those checkpoints are Llama-3.1-8B-Instruct and CAT-Q's are post-trained Qwen3-1.7B / 8B; CAT-Q brackets its cost at 1--8 h on eight A100-80GB GPUs from 1.7B to 14B. TWLA releases no checkpoint: its row is the model its released code builds that matches its reported numbers, and its rotation objective is fitted to the weights (its text mentions the calibration set, its code does not use it).}
\label{tab:competitors}
\begin{center}
\footnotesize
\setlength{\tabcolsep}{3pt}
\begin{tabular}{@{}>{\raggedright\arraybackslash}p{2.2cm}>{\raggedright\arraybackslash}p{1.0cm}>{\raggedright\arraybackslash}p{1.3cm}>{\raggedright\arraybackslash}p{1.9cm}>{\raggedright\arraybackslash}p{1.4cm}>{\raggedright\arraybackslash}p{3.2cm}>{\raggedright\arraybackslash}p{1.2cm}@{}}
\toprule
Method & Claimed & Resident & Trains? & Calibrates? & Scale axis & Source \\
\midrule
RoleQ-T (ours) & 1.585 & 1.590 & SD of scales and norm gains & yes, C4 & role: readers per row, writers per column & -- \\
Qronos + HIP + MagR & 1.58 & 1.591 & no & yes & per-OC, fp scale and zero point & rerun \\
\pttwo{} & 1.58 & 1.835 & no & yes & per-OC $\times$ g128, scale and shift & rerun \\
QTEA & 1.70 & 2.26--2.28 & no & yes, WT2 & per-OC $\times$ g128 $\alpha,\beta$ + FP8 residual (1:4 sparse) on salient columns & rerun 8B; ckpt 14B \\
CAT-Q & 1.58 & 1.71 & learned ternarization, 60 epochs & yes, C4 ($512\times2048$) & per-row $\times$ g128 & ckpt \\
TWLA & 1.58 & 4.15--4.65 & rotation, fit to weights & yes (GPTQ) & per-(row, g128) $\alpha, \mu$, saliency masks, second plane on 26\% (code, not paper) & rerun \\
PTQTP & $2 \times 1.58$ & 3.42 & no & no & per-OC $\times$ g128, one fp16 scale per plane & paper \\
QTIP 2-bit & 2 & 2.0065 (measured) & \quipsharp{}-style fine-tune (blockwise, then end-to-end) & yes & trellis codebook (HYB), BlockLDLQ & ckpt \\
GuidedQuant on QTIP, 2-bit & 2 & $\sim$2.0 & no & yes & trellis codebook (HYB), BlockLDLQ with GuidedQuant's Hessian & ckpt \\
MXFP4 & 4.25 & 4.25 & no & yes (GPTQ) & block of 32, shared E8M0 exponent & ours \\
\bottomrule
\end{tabular}
\end{center}
\end{table}

\begin{table}[h]
\caption{\textbf{Bit-accounting derivations.} Information bits per weight, counting every term a decoder needs to produce the weight (Qwen3-8B: $d = 4096$, $d_{\mathrm{ff}} = 12288$); folded or seed-regenerable rotations cost nothing, and packing is not charged (5 trits per byte would read 1.605). QTEA's terms use its own \texttt{bit\_width} function. \pttwo{}'s $\alpha$ and $\mu$ are counted as fp16 per (row, 128), the layout of its released quantizer; it publishes no per-weight accounting. TWLA's terms are measured on the model its released code builds, the code that reproduces its reported Qwen3-8B perplexity (TWLA releases no checkpoint; its paper describes one $(\alpha, \mu)$ pair per row and a single plane, while the code, of ARB-LLM lineage, adds per-block magnitude masks with separate pairs and a second plane on the salient partition) over all 10\,368 blocks: 26.21\% of the weights carry a second plane (26.48\% on the post-trained model), and the absence of a single exact zero among 1.31\,M sampled weights shows that the mask membership is consumed at inference, since $\mu$ is added to every element inside a mask regardless of its code. Even without the membership bit, TWLA sits at 3.15--3.65.}
\label{tab:bits}
\begin{center}
\footnotesize
\setlength{\tabcolsep}{3pt}
\begin{tabular}{@{}>{\raggedright\arraybackslash}p{3.1cm}>{\raggedright\arraybackslash}p{5.0cm}r>{\raggedright\arraybackslash}p{1.3cm}r@{}}
\toprule
Method & Term & bpw & Resident? & Total \\
\midrule
RoleQ-T & ternary codes, $\log_2 3$ & 1.585 & yes & \\
 & one fp16 scale per unit, $16/d$ & 0.0039 & yes & \\
 & reintroduced norm gains & 0.0007 & yes & \\
 & $r_4$ sign vector per layer & $<0.0001$ & yes & \\
 & $r_1$ / $s_1$ / $s_2$, folded into the weights & 0 & no & \textbf{1.590} \\
RoleQ-T-g128 / per-OC g128 & + ternary group scales, $g = 128$ & 0.125 & yes & \textbf{1.710} \\
per-OC g128 + shift & + per-group shift (\pttwo{}'s count) & 0.125 & yes & \textbf{1.835} \\
RoleQ-T + planes & + second-plane codes (8B budget) & 0.5658 & yes & \\
 & + one fp16 scale per planed unit & 0.0014 & yes & \\
 & + unit membership, 1 bit per unit & 0.0002 & yes & \textbf{2.157} \\
INT2-g128 (four-level) & $2 + (16 + 4)/128$ & & yes & \textbf{2.156} \\
INT3sym / INT4sym & $\log_2 7$, $\log_2 15$, $+ 16/d$ & & yes & \textbf{2.811} / \textbf{3.911} \\
MXFP4 & $4 + 8/32$ & & yes & \textbf{4.250} \\
\midrule
Qronos + HIP + MagR & three-level asymmetric codes + fp scale and zero per channel & 1.585 + 0.006 & yes & \textbf{1.591} \\
CAT-Q & ternary codes + fp16 scale per 128 & 1.585 + 0.125 & yes & \textbf{1.710} \\
\pttwo{} & ternary codes + fp16 $\alpha$ and $\mu$ per 128 & 1.585 + 0.250 & yes & \textbf{1.835} \\
PTQTP & two full ternary planes, $2 \log_2 3$ & 3.170 & yes & \\
 & fp16 $\alpha$ per plane per (row, 128), $2 \cdot 16/128$ & 0.250 & yes & \textbf{3.420} \\
QTEA, 8B / 14B & code payload & 1.781 / 1.756 & yes & \\
 & fp32 $\alpha + \beta$ per 128 & 0.500 / 0.501 & yes & \textbf{2.281} / \textbf{2.257} \\
TWLA & codes: $0.262 \cdot 2 \log_2 3 + 0.738 \log_2 3$ & 2.000 & yes & \\
 & per-(row, group) $\alpha$ and $\mu$, fp16, 4--6 pairs & 1.000--1.500 & yes & \\
 & mask membership, 1 bit per weight & 1.000 & yes & \\
 & Kronecker $L$ / $R$, learned per layer, as stored by the released code (63.7\,M parameters) & 0.147 & yes & \textbf{4.15--4.65} \\
 & \textit{codes + rotation only, the often-quoted figure} & \textit{2.147} & \textit{no weight} & \\
\bottomrule
\end{tabular}
\end{center}
\end{table}

\section{Setup Details}
\label{app:setup}

Table~\ref{tab:setup} lists the models, corpora, seeds, solver settings and evaluation protocol shared by every experiment in the paper.
\begin{table}[h]
\caption{\textbf{Models, corpora, seeds and evaluation.} Full-precision references are measured on our harness.}
\label{tab:setup}
\begin{center}
\small
\begin{tabular}{@{}l>{\raggedright\arraybackslash}p{9.8cm}@{}}
\toprule
Models & Qwen3-0.6B / 1.7B / 8B / 14B-Base; Qwen3-1.7B / 8B post-trained; Llama-3.1-8B-Base; Llama-3.1-8B-Instruct (Table~\ref{tab:llama}) \\
Calibration, paper draw & 512 C4 sequences $\times$ 512 tokens; standard draw $128 \times 2048$ for rows that match a competitor's protocol and for Llama-3.1-8B-Instruct; post-trained models: 272 chat sequences $\times$ up to 1536 tokens, half UltraChat, half self-generated \\
Calibration seeds & 45 and 46 on every cell of ours; 47 and 48 added where a cell says four seeds \\
Solver & GPTQ with act-order, damping 0.01 of the mean Hessian diagonal, lazy block 128; exact per-vector ternary step; reader refit in the $\Sigma$ metric \\
Rotation & recipe: $s_1$(l4) $+ s_2$(opt) $+ r_4$ (block 1024); ablations and placement controls: random Hadamard $r_1$ \\
Scale distillation (SD) & 400 steps, batch 8, learning rate $3 \times 10^{-3}$, KL to the cached fp teacher, 3,200 RedPajama-1T-Sample windows of 512 tokens; scales and norm gains only, codes frozen; 2\,h at 8B on one A100 \\
Perplexity & WikiText2 test at 2048 tokens \\
Zero-shot & PIQA, ARC-Easy, ARC-Challenge, HellaSwag, OpenBookQA (\texttt{acc\_norm}), WinoGrande, BoolQ (\texttt{acc}); lm-eval-harness 0.4.13 \\
MMLU & five-shot, lm-eval-harness 0.4.13 \\
Chance levels & zero-shot average 35.7, MMLU 25.0 \\
Seed statistics & mean of two seeds with the half-range as subscript\\
\midrule
\multicolumn{2}{@{}l}{fp16 references: WikiText2 ppl / zero-shot / MMLU} \\
Qwen3-0.6B-Base & 12.66 / 54.68 / 52.56 \\
Qwen3-1.7B-Base & 9.39 / 62.62 / 62.70 \\
Qwen3-8B-Base & 6.99 / 70.42 / 76.82 \\
Qwen3-14B-Base & 6.38 / 73.39 / 80.66 \\
Qwen3-1.7B post-trained & 16.67 / 60.17 / 60.30 \\
Qwen3-8B post-trained & 9.72 / 69.36 / 74.93 \\
Llama-3.1-8B-Base & 6.30 / 71.59 / 65.67 \\
Llama-3.1-8B-Instruct & 7.23 / 71.50 / 68.30 \\
\bottomrule
\end{tabular}
\end{center}
\end{table}

\section{The Placement Rule: Per-Seed Cells and Controls}
\label{app:axis}

\paragraph{Controls.}
Four controls separate the placement effect from what it could be confounded with. \emph{The code and the scale count.} On the four-level g128 grid, which changes both the code and the number of scales, the ordering RoleQ $<$ per-OC $<$ per-IC reproduces at every Qwen size, including 14B (10.76, 10.90, 11.64). At 1.7B both halves of the margin, the writer half (RoleQ against per-OC, which differ only on writers) and the reader half (RoleQ against per-IC, which differ only on readers), lie well outside the seed spread (19\% and 30\%); at 14B only the reader half does (8.2\% at $2.9\times$ the seed range, against 1.3\% at $0.5\times$ for the writer half). On Llama-8B the role and per-OC placements of the same grid tie in the recipe basis (Table~\ref{tab:equal-bits}), so on the four-level grid the ordering is established for Qwen only. \emph{The fit.} Two different ternary fits give the same ordering within noise. \emph{The rotation.} AdaDim's per-layer search agrees with the placement rule on 96.4\% of the cells in the rotated basis and on 14.3\% unrotated (Table~\ref{tab:adadim}): the axis the rule fixes is the one a reconstruction-error search prefers only after rotation, which is why we state it as a rotated-basis rule (Section~\ref{sec:rotations}). \emph{The reference placement.} The best uniform placement changes with size and family (per-IC at Qwen-8B, per-OC at 14B and on Llama), so every margin in the paper is reported against it. Llama-8B is the only cell in which both halves separate at once (writer $1.97\times$, reader $11.52\times$), between two runs at $9.2\times$ and $18.2\times$ full precision. As a consistency check, the 8B placement runs, PTQ-only under \rone{} and carried out separately from the recipe ladder, reproduce the corresponding bodies seed by seed (RoleQ 30.7 / 31.5 against 31.0 / 31.0, the \rone{} PTQ row of Table~\ref{tab:recipe-ladder}; per-OC 36.0 / 35.0 against 35.4 / 34.9).

\begin{table}[h]
\caption{\textbf{The placement rule: the cells behind Table~\ref{tab:placement}.} (a) Ternary, calibrated, \rone{}, PTQ-only: two-seed means with half-ranges (four seeds for RoleQ and per-IC at Qwen-8B), the writer half (per-OC over RoleQ), the reader half (per-IC over RoleQ), the best uniform placement, and RoleQ's lead over it with the gap as a multiple of the largest seed range. Llama-8B is the one cell in which both halves separate at once, between runs at $9\times$ and $18\times$ full precision. (b) Two grids omitted from the main table: the per-vector four-level grid (2.01 bpw), whose lead closes and reverses by 8B as it does at the integer widths, and INT8sym, where the axis makes no measurable difference (seed 45). (c) Zero-shot and MMLU scores of the integer grids where they were run and of the 14B ternary triplet (two seeds; lead: RoleQ compared to best uniform placement, \res marks notable difference; the 14B per-OC and per-IC MMLU are at chance, \dg{}).
}
\label{tab:axis-seeds}
\begin{center}
\footnotesize
\setlength{\tabcolsep}{2.5pt}
\begin{tabular}{@{}lrrrrr@{}}
\toprule
(a) ternary & 0.6B & 1.7B & Qwen-8B & 14B & Llama-8B \\
\midrule
RoleQ & \sr{351.9}{21.8} & \sr{136.2}{8.8} & \sr{31.28}{0.77}$^{4}$ & \sr{25.25}{0.34} & \sr{58.13}{2.29} \\
per-OC & \sr{387.2}{34.0} & \sr{222.3}{8.0} & \sr{35.47}{0.52} & \sr{37.89}{0.65} & \sr{114.4}{8.6} \\
per-IC & \sr{1187}{226} & \sr{305.0}{19.8} & \sr{33.39}{1.70}$^{4}$ & \sr{42.60}{0.48} & \sr{669.9}{28.1} \\
writer half & 1.10 & 1.63 & 1.14 & 1.50 & 1.97 \\
reader half & 3.37 & 2.24 & 1.07 & 1.69 & 11.5 \\
best uniform & per-OC & per-OC & per-IC & per-OC & per-OC \\
RoleQ lead & 1.10 & 1.63\res & 1.07 & 1.50\res & 1.97\res \\
gap $\div$ range & $0.6\times$ & $4.9\times$ & $0.7\times$ & $15\times$ & $6.4\times$ \\
\midrule
(b) & 0.6B & 1.7B & 8B & & \\
\midrule
four-level per-vector, RoleQ & \sr{122.1}{4.0} & \sr{52.16}{5.28} & \sr{17.61}{0.04} & & \\
\quad per-OC & \sr{129.4}{6.1} & \sr{97.38}{2.04} & \sr{16.61}{0.25} & & \\
\quad per-IC & \sr{176.0}{10.9} & \sr{86.72}{3.28} & \sr{17.21}{0.30} & & \\
\quad RoleQ lead & 1.06 & 1.66 & 0.94 & & \\
INT8sym, RoleQ & 12.669 & 9.389 & \na & & \\
\quad per-OC & 12.670 & 9.391 & \na & & \\
\quad per-IC & 12.679 & 9.394 & \na & & \\
\quad RoleQ lead & 1.0001 & 1.0002 & & & \\
\bottomrule
\end{tabular}

\medskip
\scriptsize
\setlength{\tabcolsep}{1pt}
\begin{tabular}{@{}lrrrrrrrrrr@{}}
\toprule
(c) & \multicolumn{2}{c}{0.6B INT3} & \multicolumn{2}{c}{1.7B INT3} & \multicolumn{2}{c}{1.7B INT4} & \multicolumn{2}{c}{8B INT3} & \multicolumn{2}{c}{14B ternary} \\
\cmidrule(lr){2-3} \cmidrule(lr){4-5} \cmidrule(lr){6-7} \cmidrule(lr){8-9} \cmidrule(lr){10-11}
 & zero-shot & MMLU & zero-shot & MMLU & zero-shot & MMLU & zero-shot & MMLU & zero-shot & MMLU \\
\midrule
RoleQ & \sr{48.61}{0.18} & \sr{36.10}{0.48} & \sr{59.30}{0.15} & \sr{52.01}{0.36} & \sr{62.30}{0.02} & \sr{60.61}{0.10} & \sr{69.32}{0.13} & \sr{70.71}{0.16} & \sr{52.30}{0.06} & \sr{38.52}{1.43} \\
per-OC & \sr{49.70}{0.08} & \sr{35.33}{0.14} & \sr{45.22}{0.09} & \sr{29.69}{0.06} & \sr{60.47}{0.04} & \sr{53.18}{0.13} & \sr{69.05}{0.07} & \sr{70.47}{0.17} & \sr{45.58}{0.09} & \sr{25.39}{0.34}\dg \\
per-IC & \sr{47.57}{0.34} & \sr{29.96}{0.33} & \sr{48.68}{0.41} & \sr{40.91}{1.41} & \sr{61.28}{0.45} & \sr{58.60}{0.01} & \sr{69.51}{0.15} & \sr{69.20}{0.48} & \sr{41.86}{0.49} & \sr{25.31}{0.02}\dg \\
RoleQ lead & $-1.08$\res & $+0.77$ & $+10.62$\res & $+11.10$\res & $+1.02$ & $+2.01$\res & $-0.19$ & $+0.24$ & $+6.71$\res & $+13.13$\res \\
\bottomrule
\end{tabular}
\end{center}
\end{table}

\begin{table}[h]
\caption{\textbf{AdaDim's per-layer search against the placement rule.} Left: at the GPTQ stage on Qwen3-1.7B, in 16 sampled layers, how many matrices of each kind AdaDim's reconstruction-error argmin sends to the input axis, rotated (\rone{}) and unrotated; RoleQ fixes readers to the output axis and writers to the input axis without a search. Right: the cost and the return of the search; AdaDim's own search quantizes each matrix twice with RTN before one GPTQ pass, and the ``half the solver work'' of Section~\ref{sec:exp-axis} is against a two-way GPTQ search. We did not run AdaDim end to end: it would differ from RoleQ in 4 of 112 cells at 1.7B and 6 of 112 at 8B, worth 0.003\% and 0.023\% of $\mathbf H$-metric error, three orders of magnitude below the seed noise.}
\label{tab:adadim}
\begin{center}
\small
\begin{tabular}{@{}llrr@{}}
\toprule
Kind & RoleQ & Picks IC, rotated & Picks IC, unrotated \\
\midrule
$W_{\mathrm{in}}$ & OC & 1 / 16 & 15 / 16 \\
$W_{\mathrm{gate}}$ & OC & 0 / 16 & 14 / 16 \\
$W_Q$ & OC & 0 / 16 & 11 / 16 \\
$W_K$ & OC & 0 / 16 & 15 / 16 \\
$W_V$ & OC & 3 / 16 & 16 / 16 \\
$W_{\mathrm{out}}$ & IC & 16 / 16 & 6 / 16 \\
$W_O$ & IC & 16 / 16 & 1 / 16 \\
\midrule
agreement with the rule & & 108 / 112 (96.4\%) & 16 / 112 (14.3\%) \\
\bottomrule
\end{tabular}
\qquad
\begin{tabular}{@{}lll@{}}
\toprule
 & RoleQ & AdaDim \\
\midrule
solver work per matrix & $1\times$ & $2\times$ \\
calibration data to decide & none & required \\
stored state & none & an axis flag per layer \\
$\mathbf H$-metric error, 1.7B & -- & $-0.0027$\% \\
$\mathbf H$-metric error, 8B & -- & $-0.0227$\% \\
cells that differ, 1.7B / 8B & -- & 4 / 112, 6 / 112 \\
\bottomrule
\end{tabular}
\end{center}
\end{table}

\section{The Conversion Recipe: Details}
\label{app:recipe}

\paragraph{Solver settings.}
The GPTQ sweep runs in activation order (the act-order option of the reference implementation; the paper itself uses the natural column order) through the Cholesky factor of $\mathbf{H}^{-1}$, with damping of $0.01\times$ the mean diagonal and lazy batching over blocks of 128 columns, which changes the speed but not the result. Each $\mathbf{H}$ is captured from the rotated model layer by layer, with the preceding layers already quantized. Attention readers are fitted in the LN-input metric and $W_O$ in $\E[\vz\vz^\top]$, where $\vz$ is the attention output it consumes.

\paragraph{Derivations.}
The closed forms of Section~\ref{sec:solver} follow from the structure of the objective. \emph{The exact ternary step (Eq.~\ref{eq:ternary}).} Fix a support $S$ and codes $\vt$ that vanish off $S$. Minimising $\|\vw - a\vt\|^2$ over $a$ gives $a = \vt^\top\vw/|S|$ and leaves the error $\|\vw\|^2 - (\vt^\top\vw)^2/|S|$, which is smallest when every code carries the sign of its weight; hence $a^\star(S)$ is the mean magnitude on $S$ and the error is $\|\vw\|^2 - (\sum_{i\in S}|w_i|)^2/|S|$. Among supports of a given size $k$ the sum is largest on the $k$ largest magnitudes, so the optimal support is a prefix of $|\vw|$ sorted in descending order, and the joint optimum over $(a, \vt)$ is one sort and a scan over $k$, in $O(n\log n)$: \citet{yin2019quantization} give the sort-and-cumulative-sum formula for a power-of-two scale, and \citet{hou2018lossaware} the exact solver for a real scale under a diagonal-Hessian weighting, which reduces to this objective when the weighting is uniform.
TWN's threshold~\citep{li2016twn} is the approximation to $S^\star$ that a Gaussian or uniform assumption on the weights yields. The same scan over an $L$-level grid places the integer grids exactly, where LeanQuant and NeUQI search over candidate grids~\citep{zhang2025leanquant,lin2026neuqi}. \emph{The writer step.} When the sweep reaches a writer vector, input coordinate $j$, it quantizes the compensated vector $\vw_j'$ and applies the optimal-brain-surgeon update to the coordinates not yet quantized~\citep{hassibi1993obs,frantar2022obc,frantar2023gptq}. The increase in the objective is $\|\vw_j' - \hat{\vw}_j\|^2 / [\mathbf H_F^{-1}]_{jj}$, with $\mathbf H_F$ the Hessian restricted to the unquantized coordinates, because all $m$ output rows share $\mathbf H$ and the denominator is a scalar. The quantization of $\vw_j'$ that minimises this increase is therefore the unweighted exact step of Eq.~\ref{eq:ternary} applied to $\vw_j'$, jointly over its scale and its codes. The exactness is per step; the order of the steps remains GPTQ's greedy activation order. \emph{The reader refit.} For a reader the objective is $\sum_i (\vw_i - a_i\vt_i)^\top \mathbf H (\vw_i - a_i\vt_i)$, a sum over rows with no coupling between them. The sweep rounds each row onto the grid $\{-a_i, 0, a_i\}$, with $a_i$ from the data-free exact step, compensating along the row. At the realized codes the objective is a quadratic in the scalar $a_i$, so its minimiser $a_i^\star = \vt_i^\top\mathbf H\vw_i / \vt_i^\top\mathbf H\vt_i$ is closed-form, and since the scale the sweep used is a feasible point of the same problem, the refit can only lower the error; with $\mathbf H = \mathbf I$ it reduces to the mean magnitude of Eq.~\ref{eq:ternary}. Two planes on a reader are the same argument with two scales, a $2\times 2$ system per row, and a writer has no such decoupled refit because $\mathbf H$ acts on the axis its scales index (Appendix~\ref{app:planes}).

\paragraph{Scale distillation protocol.}
The stage has two hyper-parameters, the number of steps $T$ and the batch size $B$; every window is seen once, so together they fix the number of documents at $T \cdot B$. We use $T = 400$ and $B = 8$ at a learning rate of $3\times 10^{-3}$, drawing from RedPajama-1T-Sample, the corpus on which \quipsharp{}, AQLM and EfficientQAT fine-tune~\citep{tseng2024quipsharp,egiazarian2024aqlm,chen2025efficientqat}, while calibration stays on C4. We fix 400 steps because the returns flatten quickly: at 8B the 1.59-bpw body improves from 31.0 to 12.9 perplexity in the first 100 steps and from 12.4 to 12.0 in the last 200. The teacher is never run during training; its final residuals are cached once during quantization and unembedded with the frozen head, which lets the stage fit beside the student on a single GPU.

\paragraph{Protocol notes.}
Our one deviation from the standard protocol is the shape of the calibration draw ($512\times512$ rather than $128\times2048$), which favours the baselines (Table~\ref{tab:draw}); every competitor number was produced in our harness and validated against the method's own evaluation code before being quoted, and against its paper where it reports the same model (QTEA's 8B and 14B rows match its paper exactly; TWLA's rerun is set beside its paper's numbers in Table~\ref{tab:ternary}). The ordering against CAT-Q also holds on a chat corpus (UltraChat, the post-trained model's own domain), where our run sits at $1.03\times$ and CAT-Q at $1.27\times$ of the respective full-precision perplexity.

\paragraph{Group scales.}
Ternary group scales ($g = 128$, +0.12 bpw) improve the body at either placement (Table~\ref{tab:recipe-ladder}). With the groups per output row, the standard layout, the run scores 10.87 / 61.93 / 51.33 after scale distillation; with the same groups on the role axis (RoleQ-T-g128) it scores 10.82 / 61.81 / 50.89, within $\pm 3$\% on every instrument and with the sign flipping between seeds on two of them. At PTQ the role-axis variant is ahead on all three instruments (18.49 against 19.16 perplexity, +1.2 zero-shot, +2.6 MMLU), by less than $3\times$ the seed range. With group scales, then, the role axis is on par with the standard placement rather than ahead of it.

\begin{table}[h]
\caption{\textbf{The 8B recipe ladder in both bases.} Every run of ours behind Table~\ref{tab:ternary}: Qwen3-8B-Base, C4 $512\times512$, two seeds, half-range as subscript. The \rone{} rows are the ablation that measures what the recipe's rotation is worth; the recipe rows are the numbers presented in the main text.}
\label{tab:recipe-ladder}
\begin{center}
\footnotesize
\setlength{\tabcolsep}{3pt}
\begin{tabular}{@{}>{\raggedright\arraybackslash}p{3.9cm}llrrrr@{}}
\toprule
Run & Basis & bpw & Stage & WT2 ppl$\,\downarrow$ & Zero-shot$\,\uparrow$ & MMLU$\,\uparrow$ \\
\midrule
ternary body on the role axis & \rone & 1.590 & PTQ & \sr{30.99}{0.01} & \sr{49.13}{0.31} & \sr{32.82}{0.21} \\
 & \rone & 1.590 & +SD & \sr{12.07}{0.04} & \sr{59.40}{0.35} & \sr{45.88}{0.87} \\
\quad = RoleQ-T & recipe & 1.590 & PTQ & \sr{22.42}{0.22} & \sr{50.39}{0.09} & \sr{33.32}{1.48} \\
 & recipe & 1.590 & +SD & \sr{11.33}{0.06} & \sr{60.33}{0.27} & \sr{47.59}{2.72} \\
+ ternary group scales g128, per-OC & recipe & 1.710 & PTQ & \sr{19.16}{0.07} & \sr{52.37}{0.28} & \sr{37.25}{1.68} \\
 & recipe & 1.710 & +SD & \sr{10.87}{0.06} & \sr{61.93}{0.31} & \sr{51.33}{0.38} \\
+ group scales on the role axis = RoleQ-T-g128 & recipe & 1.710 & PTQ & \sr{18.49}{0.22} & \sr{53.61}{0.28} & \sr{39.85}{0.66} \\
 & recipe & 1.710 & +SD & \sr{10.82}{0.02} & \sr{61.81}{0.19} & \sr{50.89}{0.69} \\
ternary body + text-Fisher planes & \rone & 2.157 & PTQ & \sr{12.83}{0.05} & \sr{57.57}{0.16} & \sr{53.99}{0.13} \\
 & \rone & 2.157 & +SD & \sr{9.73}{0.02} & \sr{64.81}{0.06} & \sr{63.19}{0.34} \\
\quad = RoleQ-T + planes & recipe & 2.157 & PTQ & \sr{11.32}{0.01} & \sr{59.51}{0.30} & \sr{56.04}{0.03} \\
 & recipe & 2.157 & +SD & \sr{9.30}{0.02} & \sr{65.89}{0.34} & \sr{62.75}{0.41} \\
\bottomrule
\end{tabular}
\end{center}
\end{table}

\begin{table}[h]
\caption{\textbf{Llama-3.1-8B-Instruct at 2 bpw against the trellis-coded class.} Scores and, in parentheses, the share of the above-chance headroom retained (fp16: 7.23 ppl, 71.50 zero-shot, 68.30 MMLU; chance 35.7 and 25.0). QTIP (whose stage includes its \quipsharp{}-style fine-tune) and GuidedQuant are their released checkpoints re-scored on our harness (QTIP's WikiText2 reproduces its published value to 0.001, in its own convention on the materialised model). Our plane budget is set so that the rate matches QTIP-2bit to three decimals (1.590 + 0.4165). Calibration on the standard draw $128 \times 2048$. The recipe rows are two-seed on every instrument; the \rone{} rows (seed 45) are the same method in the ablation basis. This is a negative result: at matched bits the trellis-coded methods lead on every instrument ($-4.3$ zero-shot and $-8.9$ MMLU against QTIP after distillation, $-2.5$ and $-6.6$ against GuidedQuant), and the change of basis narrows the gap (from $-5.0$ / $-11.9$ in \rone{}) without closing it.}
\label{tab:llama}
\begin{center}
\footnotesize
\setlength{\tabcolsep}{2pt}
\begin{tabular}{@{}llrrrr@{}}
\toprule
Method & Stage & bpw & WT2 ppl ($\times$ fp) & Zero-shot (\%) & MMLU (\%) \\
\midrule
QTIP 2-bit & PTQ + FT & 2.0065 & 8.83 (1.22) & 68.13 (90.6) & 59.93 (80.7) \\
GuidedQuant on QTIP, 2-bit & PTQ & $\sim$2.0 & 10.11 (1.40) & 66.29 (85.4) & 57.58 (75.2) \\
RoleQ-T + planes, recipe & PTQ & 2.0065 & \sr{21.58}{0.86} (2.98) & \sr{57.86}{0.31} (61.9) & \sr{40.50}{0.26} (35.8) \\
RoleQ-T + planes, recipe & +SD & 2.0065 & \sr{11.06}{0.06} (1.53) & \sr{63.81}{0.02} (78.5) & \sr{51.02}{0.63} (60.1) \\
RoleQ-T + planes, \rone & PTQ & 2.0065 & 26.34 (3.64) & 55.75 (56.0) & 37.62 (29.1) \\
RoleQ-T + planes, \rone & +SD & 2.0065 & 11.50 (1.59) & 63.09 (76.5) & 47.99 (53.1) \\
\bottomrule
\end{tabular}
\end{center}
\end{table}

\section{Sparse Planes: Details}
\label{app:planes}

\paragraph{Refit by role.}
For readers the two scales of a unit with a second plane are re-solved jointly by an exact $2\times 2$ weighted least squares in the $\Sigma$ metric, which decouples per unit because $\Sigma$ acts on the input axis. For writers there is no joint refit: $\mathbf H$ acts on the axis the scales index, so the exact refit is one coupled $2n\times 2n$ system and a per-unit shortcut would cancel $\mathbf H$ out; the coupling is handled inside the GPTQ recursion instead.

\paragraph{Probe protocol.}
The scored span is the chat-rendered answer tokens of a knowledge item (end marker included, as the evaluation scores it) or the second half of each 384-token C4 window; prompt tokens serve as context and are never scored. The Fisher is accumulated in fp32 over every quantized weight, one forward-backward pass per item over 720 items, and kept in two interleaved item halves for the split-half check (17\,GB of host memory at 1.7B; probes at 8B and above need an H200). The budget over the one-plane body targets 2.156 bpw for the Base rows ($+0.556$ / $0.562$ / $0.566$ / $0.567$ code bits at 0.6B / 1.7B / 8B / 14B and Llama-8B) and 1.994 for the post-trained rows ($+0.40$); (layer, kind) cells whose selected fraction falls below 2\% are dropped as noise, which leaves about 180--200 cells with a second plane. The score is basis-specific for $W_V$ and $W_O$ under $s_2$ and for $W_{\mathrm{out}}$ under $r_4$, so the recipe-basis planes run recomputes its saliency rather than reusing the $r_1$-basis ranking.

\paragraph{Controls.}
The four-level control in every equal-bits comparison is the exact-grid INT2-g128 fit; a min/max grid would measure grid placement inside a vector, a $20\times$ effect at 1.7B (888 against 38), rather than the format. At 8B after scale distillation the planes lead both four-level controls on perplexity by 1.6\% in the recipe basis, with the same sign on every seed pairing at $2.7\times$ the per-OC run's seed range, and lead MMLU by 2.4 points, within the seed variance (Table~\ref{tab:equal-bits}). The disattenuated correlation between the text-Fisher and knowledge-Fisher rankings is 0.93 (split-half $\rho$ 0.87--0.98), and the checkpoint on which the score is computed moves every instrument uniformly, by 4--10\%.

\begin{figure}[t]
  \centering
  \includegraphics[width=\textwidth]{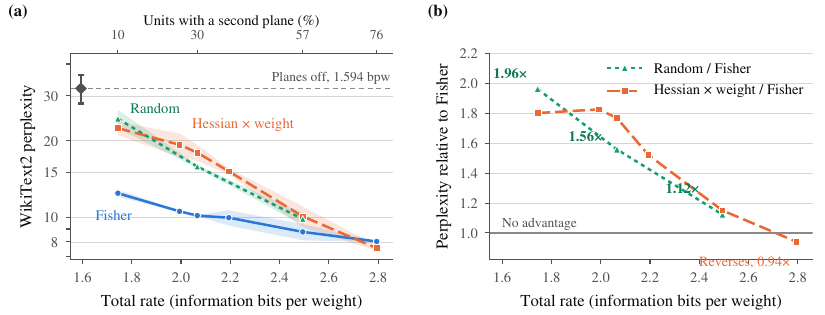}
  \caption{%
    \textbf{The plane budget sweep.}
    Qwen3-1.7B (post-trained), PTQ stage, WikiText2 perplexity, two calibration
    seeds at every budget; bands are the seed range.
    (a)~Perplexity against the total rate as the second plane is extended to
    more units (upper axis), for three placements: the per-unit Fisher saliency,
    Hessian${}\times{}$weight saliency, and random units; the planes-off run at
    $1.594$ bits is the common origin.  Random was run at three budgets.
    (b)~The same cells as a ratio to the Fisher placement: its advantage over
    random decays from $1.96\times$ at $+0.15$ bits to $1.12\times$ at $+0.90$,
    and Hessian${}\times{}$weight overtakes it at $+1.20$ bits ($0.94\times$).
    The decay is measured at 1.7B and is not claimed at 8B.}
  \label{fig:sweep}
\end{figure}

\paragraph{Budget, selectors and items.}
The budget sweep (Figure~\ref{fig:sweep}) and the full selector set (Table~\ref{tab:selectors}) are measured on the post-trained 1.7B model. Spearman correlation with realized per-unit PTQ error does not predict useful placement: Hessian$\times$weight correlates strongly ($\rho = +0.96$) yet performs worst among selectors, while the knowledge Fisher anti-correlates ($\rho = -0.47$) and performs much better. Error magnitude alone therefore appears insufficient to identify the units that benefit most from extra precision. The item set decides what is recovered, which is why the score is conditioned on it at all: on that model the text Fisher is best on perplexity and worst on factual recall, the knowledge Fisher the reverse, and the checkpoint on which the score is computed changes only the size of the effect.

\begin{table}[h]
\caption{\textbf{Matched bits in the recipe basis.} Ternary + text-Fisher planes (2.157 bpw at Qwen-8B, 2.158 at Llama-8B) against the four-level INT2-g128 grid (2.156) with scales per output channel (the standard) and on the role axis (the best four-level control); all three are ours, with the same solver, kernel and calibration; two seeds, half-range as subscript. The lead is the better control's perplexity over ours (or the point gap against the better control), with the gap as a multiple of the largest seed range among the runs compared; \res{} marks $\ge 3\times$.
On Llama the role-axis grid ties per-OC, with the sign against it. The \rone{}-basis perplexities at 0.6B and 1.7B are in Table~\ref{tab:placement}.}
\label{tab:equal-bits}
\begin{center}
\footnotesize
\setlength{\tabcolsep}{3pt}
\begin{tabular}{@{}llrrrrl@{}}
\toprule
\multicolumn{7}{@{}l}{\textbf{(a) WikiText2 perplexity}} \\
Model & Stage & planes & RoleQ-INT2-g128 & per-OC INT2-g128 & lead & gap $\div$ range \\
\midrule
Qwen-8B & PTQ & \sr{11.43}{0.02} & \sr{11.79}{0.01} & \sr{11.88}{0.03} & 1.032 & $7.5\times$\res \\
Qwen-8B & +SD & \sr{9.30}{0.02} & \sr{9.44}{0.00} & \sr{9.44}{0.03} & 1.016 & $2.7\times$ \\
Llama-8B & PTQ & \sr{13.58}{0.05} & \sr{17.42}{0.15} & \sr{17.21}{0.40} & 1.27 & $5.8\times$\res \\
Llama-8B & +SD & \sr{9.59}{0.04} & \sr{10.25}{0.04} & \sr{10.22}{0.05} & 1.065 & $6.1\times$\res \\
\midrule
\multicolumn{7}{@{}l}{\textbf{(b) zero-shot average}} \\
\midrule
Qwen-8B & PTQ & \sr{59.51}{0.30} & \sr{60.68}{0.78} & \sr{60.18}{1.00} & $-1.17$ & $0.6\times$ \\
Qwen-8B & +SD & \sr{65.89}{0.34} & \sr{65.70}{0.48} & \sr{65.73}{0.38} & $+0.16$ & $0.2\times$ \\
Llama-8B & PTQ & \sr{59.76}{0.23} & \sr{56.55}{0.63} & \sr{56.86}{0.45} & $+2.89$ & $3.2\times$\res \\
Llama-8B & +SD & \sr{64.12}{0.18} & \sr{62.67}{0.18} & \sr{62.92}{0.42} & $+1.20$ & $1.4\times$ \\
\midrule
\multicolumn{7}{@{}l}{\textbf{(c) five-shot MMLU}} \\
\midrule
Qwen-8B & PTQ & \sr{56.04}{0.03} & \sr{54.45}{0.14} & \sr{55.39}{0.43} & $+0.65$ & $0.8\times$ \\
Qwen-8B & +SD & \sr{62.75}{0.41} & \sr{60.36}{0.53} & \sr{60.30}{0.40} & $+2.40$ & $2.3\times$ \\
Llama-8B & PTQ & \sr{43.48}{1.03} & \sr{32.52}{1.13} & \sr{32.76}{1.39} & $+10.72$ & $3.9\times$\res \\
Llama-8B & +SD & \sr{48.97}{0.47} & \sr{43.93}{0.38} & \sr{44.04}{1.00} & $+4.93$ & $2.5\times$ \\
\bottomrule
\end{tabular}
\end{center}
\end{table}

\begin{table}[h]
\caption{\textbf{Every selector at 1.7B, grouped by the family of the score.} Qwen3-1.7B post-trained, $+0.47$ bits of planes (2.066 bpw), PTQ WikiText2 perplexity and the same runs after scale distillation (fp 5.64); two seeds with the half-range as subscript, a cell without a subscript is one seed; $\rho$ as in Table~\ref{tab:saliency}. The loss-sensitivity scores lie at 9.7--10.3 and the error-magnitude scores at 13.7--17.9, with random placement among the latter; three of the four error-magnitude scores, including the solver's own measured error, fail to beat random, so a better estimate of the error would not help. The $3.3\times$ spread across selectors at PTQ narrows to $1.22\times$ after scale distillation.}
\label{tab:selectors}
\begin{center}
\small
\setlength{\tabcolsep}{4pt}
\begin{tabular}{@{}lrrr@{}}
\toprule
Selector & $\rho$ & PTQ & after SD \\
\midrule
\multicolumn{4}{@{}l}{\textit{loss-sensitivity scores}} \\
text Fisher (C4 windows) & $-0.27$ & \sr{9.70}{0.17} & 5.45 \\
knowledge Fisher & $-0.47$ & \sr{10.15}{0.05} & 5.58 \\
TaCQ score on the knowledge items & -- & \sr{10.30}{0.19} & 5.59 \\
\multicolumn{4}{@{}l}{\textit{error-magnitude scores}} \\
activation norm & -- & 13.71 & 5.73 \\
one-plane residual magnitude & -- & 16.00 & 5.67 \\
attributed PTQ error & -- & 16.07 & 5.61 \\
Hessian$\times$weight & $+0.96$ & \sr{17.95}{1.31} & 5.72 \\
\multicolumn{4}{@{}l}{\textit{controls}} \\
random units & $0.00$ & \sr{15.8}{0.2} & 5.75 \\
planes off (1.594 bpw) & -- & \sr{32.1}{4.1} & 6.64 \\
\bottomrule
\end{tabular}
\end{center}
\end{table}

\section{Rotation Details}
\label{app:rotations}

\paragraph{Slots.}
$r_1$ rotates the residual basis and folds into every reader's input and every writer's output; $r_2$ mixes V$\to$O per kv-head, acting on $W_V$'s columns and $W_O$'s rows; $r_3$, the per-head Q/K dimensions, is offline only without RoPE and unused here; $r_4$ rotates the MLP intermediate dimension, $W_{\mathrm{out}}$'s input space, as a runtime Hadamard hook plus fold. The names $r_1$--$r_4$ are SpinQuant's~\citep{liu2025spinquant}; the positions are QuaRot's~\citep{ashkboos2024quarot}. The learned $s_1$ and $s_2$ refine the $r_1$ and $r_2$ slots at the same endpoints, fitted to the weights alone with no labels, so the recipe stays data-free up to this point.

\paragraph{Basis labelling.}
Because the basis has a significant impact on the quantization results, we treat it as part of a run's identity: every run quantized in the recipe basis $s_1(\mathrm{l4}) + s_2(\mathrm{opt}) + r_4$ is labelled as such, runs from the two bases never share a table row, and no plane effect is computed across them. $r_4$ uses a block size of 1024 at 1.7B and 8B; its runtime Hadamard is the one a W2A8 path would reuse.

\paragraph{Why the placement rule depends on the rotation.}
The structural reason given in Section~\ref{sec:rotations}, that $r_1$ mixes the residual-stream axis and leaves the unit axis untouched, makes two predictions that the appendix tests. First, the one slot that does mix a unit axis, $r_4$ on the columns of $W_{\mathrm{out}}$, should compress the writer half of the margin specifically; it does, from $1.63\times$ to $1.17\times$ at 1.7B on both seeds (Table~\ref{tab:rotation}d). Second, the axis a reconstruction-error search prefers should coincide with the rule's only after rotation; AdaDim's argmin agrees with the rule on 96.4\% of cells rotated and on 14.3\% unrotated (Table~\ref{tab:adadim}).

\section{Ablation Details}
\label{app:ablations}

\paragraph{Rotation.}
Table~\ref{tab:rotation} measures what each slot and objective is worth on 1.7B ternary. The random Hadamard $r_1$ alone takes PTQ perplexity from 1245.9 to 119.8 and, after scale distillation, from 67.5 to 23.7. Among the $s_1$ objectives, \emph{opt} reaches the lowest weight-space error ($0.65\times$ that of $r_1$) but the highest perplexity (130.3), \emph{l4} reaches $0.91\times$ and 89.99, and TWLA's tri-modal objective 158.4: the objective that fits the weights best does not fit the model best. With $s_1$ = l4 fixed, $r_4$ alone lowers PTQ perplexity by 7.6\% (89.99 to 83.17), a learned $s_2$ beats a random $r_2$ in the same slot (73.66 against 78.27), and the two $s_2$ objectives tie after scale distillation (20.26 / 20.27). The full recipe basis scores 73.66 at PTQ ($7.85\times$ full precision) and 20.26 after scale distillation at 1.7B, and 22.42 against 30.99 under $r_1$ at 8B (Table~\ref{tab:recipe-ladder}). Paired with per-OC scales, $s_1(\mathrm{opt})$ collapses to 585.6, against 130.3 with the placement rule.

\paragraph{Solver and draw.}
Table~\ref{tab:rotation}c isolates the writer step: the exact per-vector fit against TWN's threshold and nearest rounding, and the reader refit on and off. Placement runs share 82--86\% of their trits with one another and 83.8\% with a second-seed run of the same configuration, so the margin between placements is a distributional effect rather than a different set of rounding decisions. The shape of the calibration draw ($512\times512$ against $128\times2048$) moves every method by more than most format effects (Table~\ref{tab:draw}).

\begin{table}[h]
\caption{\textbf{Rotation bake-off and solver controls.} (a) Qwen3-1.7B-Base, RoleQ format, C4 $512\times512$ seed 45, WikiText2 perplexity at PTQ and after SD; the weight-space column is the ternary reconstruction error relative to the random Hadamard. The objective with the lowest weight-space error has the highest perplexity, and the learned rotation paired with per-OC scales collapses. (b) The $s_2$ and $r_4$ slots isolated, $s_1$ = l4 throughout. (c) Solver controls on the writer step: Qwen3-1.7B, minipile $256 \times 512$, \rone{}, PTQ seed 42 and 100-step SD on two seeds. (d) The $r_4$ intervention: the writer-half margin at 1.7B ternary with and without $r_4$, two seeds; $r_4$ is the one slot that mixes a unit axis, and it compresses the half that lives on the writers by two thirds in log terms.}
\label{tab:rotation}
\begin{center}
\footnotesize
\setlength{\tabcolsep}{2.5pt}
\begin{tabular}{@{}>{\raggedright\arraybackslash}p{3.7cm}llcrrr@{}}
\toprule
(a) Rotation & $s_1$ objective & slot 2 & $r_4$ & weight-space & PTQ & +SD \\
\midrule
none & -- & -- & -- & -- & 1245.9 & 67.53 \\
random Hadamard $r_1$ & -- & -- & -- & $1.00\times$ & 119.8 & 23.70 \\
$s_1$, sparsifying (opt) & opt & -- & -- & $0.65\times$ & 130.3 & 23.23 \\
$s_1$, flattening (l4) & l4 & -- & -- & $0.91\times$ & 89.99 & 21.39 \\
$s_1$, TWLA's tri-modal & trigmm & -- & -- & $1.00\times$ & 158.4 & 24.81 \\
$s_1 + s_2 + r_4$ & opt & $s_2$ & yes & $0.65\times$ & 92.81 & 21.00 \\
$s_1 + s_2 + r_4$ = the recipe & l4 & $s_2$ & yes & $0.91\times$ & 73.66 & 20.26 \\
$s_1$ (opt) with per-OC scales & opt & -- & -- & -- & 585.6 & 30.42 \\
$s_1 + s_2 + r_4$ with per-OC scales & opt & $s_2$ & yes & -- & 112.3 & 21.34 \\
\midrule
(b) Slot 2 and $r_4$ & & slot 2 & $r_4$ & weight-space (slot 2) & PTQ & +SD \\
\midrule
 & l4 & -- & -- & -- & 89.99 & 21.39 \\
 & l4 & $s_2$ (opt) & -- & $0.74\times$ & 87.59 & 20.96 \\
 & l4 & -- & yes & -- & 83.17 & 20.82 \\
 & l4 & $r_2$ random & yes & -- & 78.27 & 21.41 \\
 & l4 & $s_2$ (opt) & yes & $0.74\times$ & 73.66 & 20.26 \\
 & l4 & $s_2$ (l4) & yes & $0.88\times$ & 70.86 & 20.27 \\
\midrule
(c) Solver control, 1.7B & & & & & PTQ & SD100, two seeds \\
\midrule
exact per-vector solve + reader refit & & & & & 90.0 & 24.45 / 24.75 \\
reader refit off & & & & & 92.6 & 24.38 / 24.75 \\
TWN threshold in the writer step & & & & & 97.1 & 24.81 / 26.36 \\
nearest rounding in the writer step & & & & & 112.8 & 26.66 / 25.71 \\
per-OC, nearest, no refit (vanilla) & & & & & 138.6 & 27.46 / 28.25 \\
writers per output row, refit kept & & & & & 202.8 & \na \\
\midrule
(d) $r_4$ intervention, 1.7B ternary & & & & & $r_1$ & $r_1 + r_4$ \\
\midrule
RoleQ & & & & & \sr{136.2}{8.8} & \sr{122.5}{2.8} \\
per-OC & & & & & \sr{222.3}{8.0} & \sr{143.6}{3.9} \\
per-OC / RoleQ & & & & & 1.632 & 1.173 \\
\bottomrule
\end{tabular}
\end{center}
\end{table}

\begin{table}[h]
\caption{\textbf{Calibration draw-shape sensitivity.} PTQ WikiText2 perplexity of the same configuration calibrated on the standard draw ($128 \times 2048$ tokens) and on the paper draw ($512 \times 512$), seed 45, multiple of fp16 in parentheses. The draw shape moves every method, including \pttwo{} on its own corpus, by more than most format effects, and it favours the baselines as much as it favours us. The ternary rows are the RoleQ-T body under \rone{}.}
\label{tab:draw}
\begin{center}
\small
\begin{tabular}{@{}lrrr@{}}
\toprule
Model, method & standard $128 \times 2048$ & paper $512 \times 512$ & ratio \\
\midrule
Qwen3-8B-Base, ternary body & 33.90 (4.85) & 30.32 (4.33) & $1.12\times$ \\
Llama-3.1-8B-Base, ternary body & 127.9 (20.3) & 56.95 (9.04) & $2.25\times$ \\
Qwen3-1.7B-Base, ternary body & 282.6 (30.1) & 119.8 (12.8) & $2.36\times$ \\
\pttwo{} on Qwen3-8B-Base, WikiText2 draw & 28.96 (4.14) & 25.73 (3.68) & $1.13\times$ \\
\pttwo{} on Qwen3-8B-Base, C4 draw & 84.49 (12.1) & 67.95 (9.72) & $1.24\times$ \\
\bottomrule
\end{tabular}
\end{center}
\end{table}

\end{document}